\documentclass[10pt,twocolumn,letterpaper]{article}

\usepackage[pagenumbers]{Styles/iccv} 

\usepackage{booktabs}
\usepackage{graphicx}
\usepackage{multirow}
\usepackage{tabularx}
\usepackage{amsmath}
\usepackage{enumitem}
\usepackage{soul}

\usepackage{bbold}

\definecolor{softgold}{RGB}{255, 230, 153}  
\definecolor{pastelblue}{RGB}{189, 215, 238}  

\newcommand{\hlc}[2][yellow]{ {\sethlcolor{#1} \hl{#2}} }
\usepackage[most]{tcolorbox}

\definecolor{iccvblue}{rgb}{0.21,0.49,0.74}
\usepackage[pagebackref,breaklinks,colorlinks,allcolors=iccvblue]
{hyperref}

\def\paperID{} 
\def\confName{}
\def\confYear{}

\title{AutoComPose: Automatic Generation of Pose Transition Descriptions for \\ Composed Pose Retrieval Using Multimodal LLMs}

\author{%
Yi-Ting Shen$^{1}$\thanks{Contributed equally}~~~~~~~Sungmin Eum$^{2}$\footnotemark[1]~~~~~~~Doheon Lee$^1$~~~~~~~Rohit Shete$^1$~~~~~~~Chiao-Yi Wang$^1$ \\ \vspace{-0.9em} \\
Heesung Kwon$^2$~~~~~~~Shuvra S. Bhattacharyya$^1$ \\ \vspace{-0.9em} \\
$^1$University of Maryland, College Park~~~~~~~~~~$^2$DEVCOM Army Research Laboratory
}

\begin{document}
\maketitle

\begin{abstract}
Composed pose retrieval (CPR) enables users to search for human poses by specifying a reference pose and a transition description, but progress in this field is hindered by the scarcity and inconsistency of annotated pose transitions. Existing CPR datasets rely on costly human annotations or heuristic-based rule generation, both of which limit scalability and diversity. In this work, we introduce \textbf{AutoComPose}, the first framework that leverages multimodal large language models (MLLMs) to automatically generate rich and structured pose transition descriptions. Our method enhances annotation quality by structuring transitions into fine-grained body part movements and introducing mirrored/swapped variations, while a \textit{cyclic consistency} constraint ensures logical coherence between forward and reverse transitions. To advance CPR research, we construct and release two dedicated benchmarks, AIST-CPR and PoseFixCPR, supplementing prior datasets with enhanced attributes. Extensive experiments demonstrate that training retrieval models with AutoComPose yields superior performance over human-annotated and heuristic-based methods, significantly reducing annotation costs while improving retrieval quality. Our work pioneers the automatic annotation of pose transitions, establishing a scalable foundation for future CPR research.
\end{abstract}
\section{Introduction}
\label{sec:intro}

\begin{figure}
\centering
\includegraphics[width=\columnwidth]{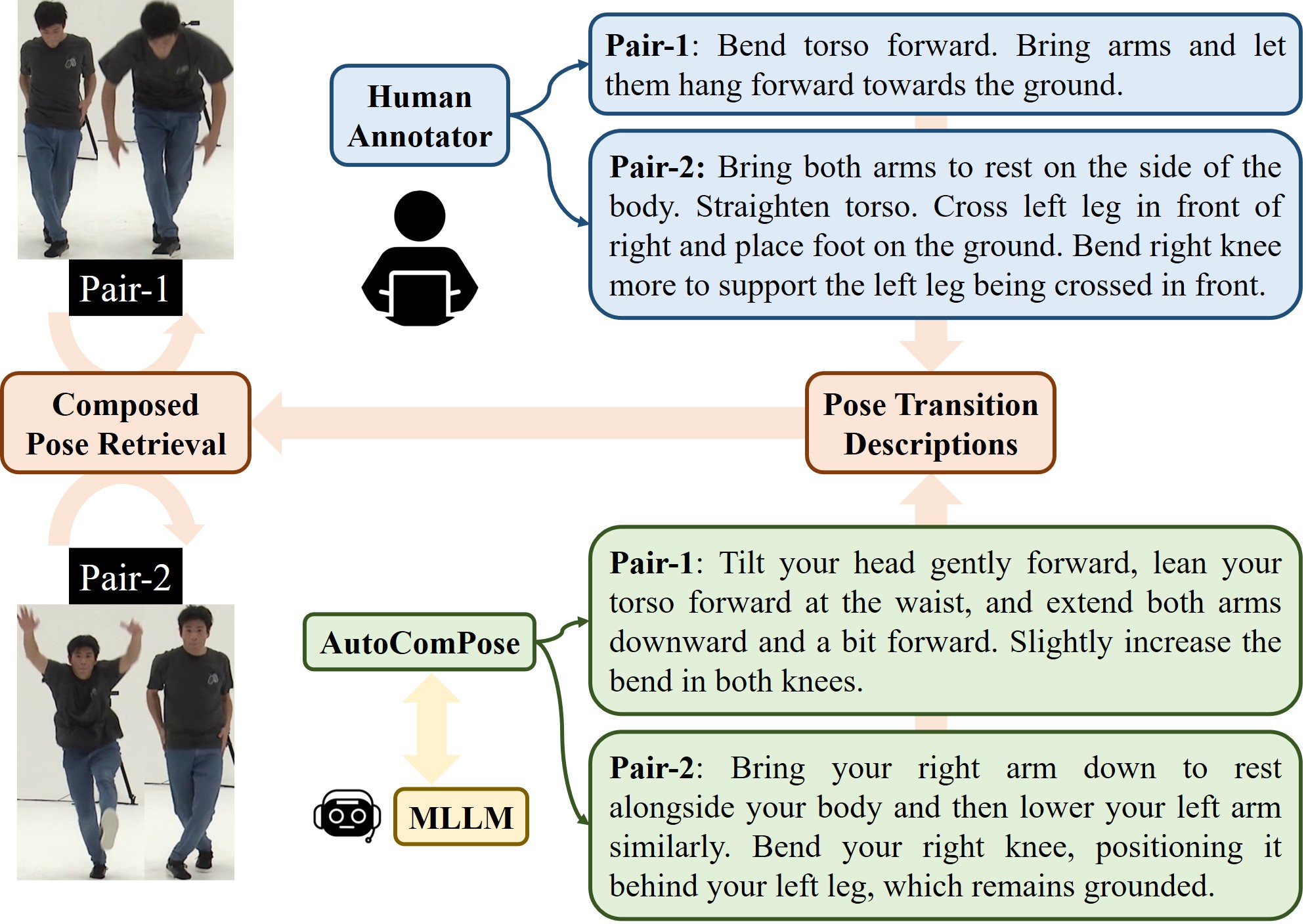}
\caption{Training composed pose retrieval models require a large and diverse set of human pose transition descriptions, demanding significant human annotation efforts. We propose AutoComPose, a framework that automatically generates pose transition descriptions using multimodal large language models (MLLMs), enhancing pose retrieval performance.}
\label{fig:cpr}
\end{figure}

Understanding and retrieving human body poses remains a fundamental challenge in computer vision, particularly in designing systems that allow users to search for specific poses through intuitive queries. These queries typically fall into two main categories: \textit{visual queries}~\cite{mori2015pose, jenicek2019linking}, where users specify their desired pose using an image, and \textit{textual queries}~\cite{delmas2022posescript, feng2024chatpose, wang2024adaptive}, which describe poses in natural language. Visual queries are intuitive but impractical when an exact reference image is unavailable. Textual queries, on the other hand, offer fine-grained control but demand precise and often tedious descriptions. This trade-off highlights a fundamental gap: how can we enable users to retrieve human poses efficiently while balancing ease of use and descriptive flexibility?

A promising solution to this gap is \textit{composed pose retrieval} (CPR)~\cite{kim2021fixmypose}, which allows users to refine a reference pose using a textual transition description. As a specialized case of composed image retrieval (CIR)~\cite{baldrati2023composed,liu2021image,wu2021fashion,saito2023pic2word,jang2024visual}, CPR leverages vision-language models (VLMs)~\cite{radford2021learning,li2022blip} trained on triplets of \{reference image, target image, transition description\}, aligning composed query features with target pose image representations. However, a key bottleneck in this pipeline is the costly and error-prone process of collecting transition descriptions, particularly for highly articulated human poses.

Unlike CIR datasets such as CIRR~\cite{liu2021image} and FashionIQ~\cite{wu2021fashion}, which focus on simple transitions like attribute modifications (e.g., color and shape, or background), spatial shifts (e.g., position and viewpoint), CPR requires fine-grained motion descriptions across multiple articulated body parts. This makes human annotation especially challenging: annotators may overlook subtle changes, use inconsistent phrasing, or introduce subjective language~\cite{kim2021fixmypose,delmas2023posefix}. The result is an expensive, inconsistent, and often insufficiently diverse dataset of pose transitions. Addressing this annotation bottleneck is essential for improving CPR performance. Despite the growing interest in CPR, no prior work has explored \textit{automatically generating pose transition descriptions} tailored for this task. We take the first step in this direction by leveraging multimodal large language models (MLLMs) to construct expressive, structured, and scalable transition annotations.

In this work, we present \textbf{AutoComPose} (short for \textbf{Auto}matic generation of \textbf{Com}posed \textbf{Pose} descriptions), an MLLM-based framework for automatic human pose transition description generation targeted for CPR, as shown in Fig.~\ref{fig:cpr}. Prior attempts to pose transition annotation rely on either human labeling~\cite{kim2021fixmypose,delmas2023posefix} or 3D human pose data with hand-crafted generation rules~\cite{delmas2022posescript,delmas2023posefix}, both of which face scalability issues and may require extra information (e.g., 3D human pose) during inference, limiting their applicability. On the other hand, recent advances in multimodal large language models (MLLMs)~\cite{wu2023multimodal,liu2024improved,wu2024next} suggest that these models possess a degree of human pose understanding~\cite{feng2024chatpose}, making them a promising tool for generating rich and diverse pose transition descriptions.

While harnessing MLLMs for generating pose transition descriptions, AutoComPose is designed to ensure expressive, diverse, and robust transition descriptions for CPR. For \textit{expressiveness}, it adopts a two-step approach inspired by human pose descriptions: it first generates detailed body part-based transitions and then integrates and paraphrases them into a complete description. For \textit{diversity}, AutoComPose introduces swapped and mirrored pose variations, enriching the training data.

However, MLLMs are known to generate incorrect or inconsistent descriptions~\cite{bai2024hallucination}, such as misassigning body part movements or fabricating transitions that do not exist in the input images. These errors can propagate into CPR training, potentially degrading retrieval performance. To improve \textit{robustness}, AutoComPose incorporates a \textit{cycle consistency loss} during training, enforcing a constraint that ensures the composed features (i.e., reference image + transition description) can be mapped back to the reference image features via the reverse transition description features. The reverse transition descriptions are obtained by prompting the MLLM with swapped input image pairs. This self-verification mechanism reduces misaligned transitions and improves reliability.

To further advance research in CPR, we introduce two new benchmarks constructed from AIST++~\cite{li2021ai} and PoseFix~\cite{delmas2023posefix}, namely \textit{AIST-CPR} and \textit{PoseFixCPR}, respectively, to supplement a previous benchmark~\cite{kim2021fixmypose}. These benchmarks establish a foundation for evaluating CPR models under diverse motion scenarios, facilitating future progress in the field. We conduct comprehensive experiments comparing AutoComPose-generated descriptions against human annotations and rule-based methods~\cite{delmas2022posescript,delmas2023posefix} across all benchmarks. Results show that models trained with AutoComPose consistently yields competitive or superior performance across multiple architectures with respect to the baselines. By significantly reducing annotation costs while maintaining high-quality descriptions, AutoComPose paves the way for more scalable and effective CPR solutions.\smallskip

\noindent{Our key contributions are as follows:}
\begin{itemize}
    \item \textbf{AutoComPose:} We propose the first MLLM-based, scalable framework for \textit{automatically} generating pose transition descriptions.
    \item \textbf{New direction and benchmarks:} We initiate the study of auto-annotations for CPR and release two new CPR benchmarks for standardized evaluation.
    \item \textbf{Expressive and diverse descriptions:} We produced fine-grained body part-based transitions while enhancing diversity through swap/mirror/paraphrase augmentations.
    \item \textbf{Cyclic consistency loss:} We introduce a novel cyclic loss to mitigate MLLM errors by enforcing semantic consistency across bi-directional transitions.
    \item \textbf{State-of-the-art performance:} Evaluations on FixMyPose, AIST-CPR, and PoseFixCPR show that AutoComPose outperforms human-annotations and existing rule-based baselines by a large margin.
\end{itemize}
\section{Related Work}
\label{sec:related}

\begin{figure*}
\centering
\includegraphics[width=\linewidth]{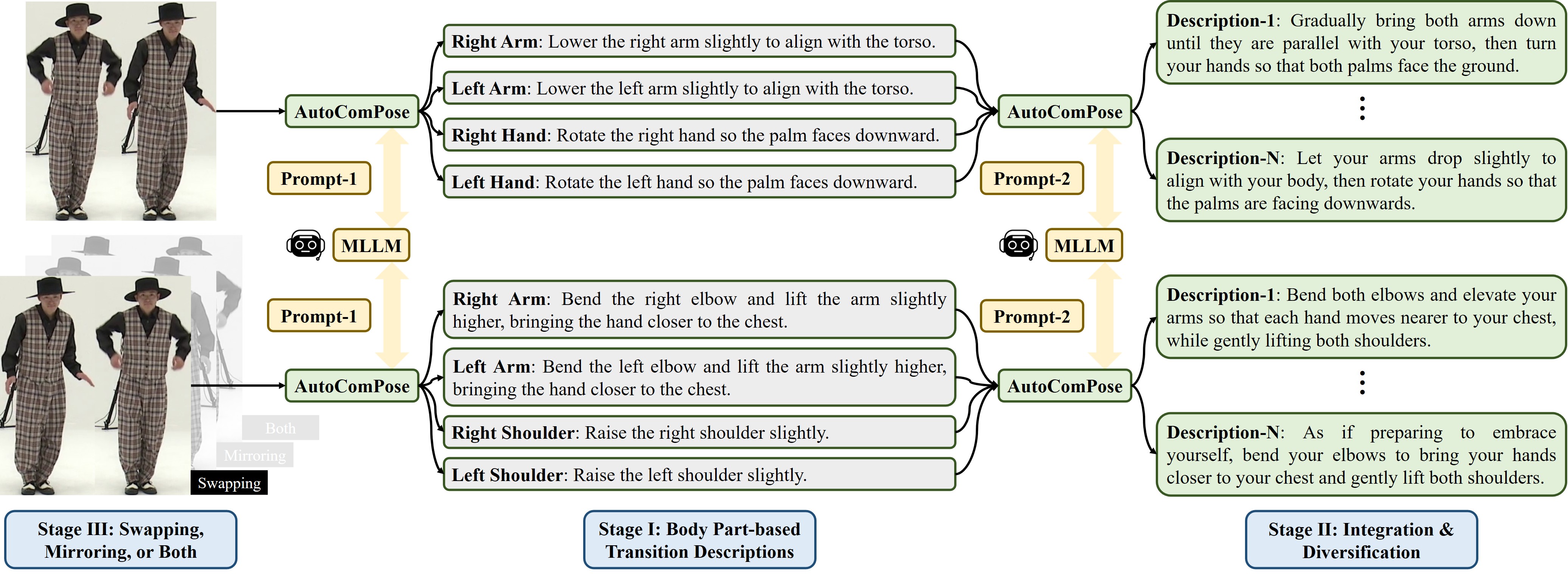}
\caption{AutoComPose generates pose transition descriptions through a three-stage process: In {\bf Stage \uppercase\expandafter{\romannumeral 1}}, it extracts body part-based transition descriptions as an intermediate representation for human body movements to enhance expressiveness. In {\bf Stage \uppercase\expandafter{\romannumeral 2}}, it integrates these body part-based descriptions and diversifies them through paraphrasing. Finally, in {\bf Stage \uppercase\expandafter{\romannumeral 3}}, it generates descriptions for input pose pairs after applying swapping, mirroring, both transformations to further improve diversity.}
\label{fig:AutoComPose}
\end{figure*}

\noindent{\bf Pose Retrieval.} Pose retrieval identifies images corresponding to specific human poses. Traditional methods~\cite{ferrari2009pose, mori2015pose} relied on handcrafted features and explicit pose representations. Deep learning has since enabled learning robust feature embeddings for improved retrieval. Wang et al.~\cite{wang2018learning} introduced a framework for discriminative local descriptors of 3D surface shapes, enhancing pose retrieval accuracy. Pavlakos et al.~\cite{pavlakos2018ordinal} leveraged ordinal depth supervision to address depth ambiguity in 3D pose estimation. Our work extends beyond pose retrieval by generating natural language descriptions of pose transitions, enriching retrieval with semantic context.\smallskip

\noindent{\bf Composed Image Retrieval.} Composed Image Retrieval (CIR) enables retrieving images based on a reference image and a textual modification. Recent advances leverage vision-language models (VLMs)~\cite{wen2024fusion, baldrati2023zero, zhao2024neucore, song2024survey} with transformer-based encoders, contrastive learning, and zero-shot methods. Wen et al.~\cite{wen2024fusion} proposed a multimodal fusion strategy optimizing linguistic and visual representations. Zhao and Xu introduced NEUCORE~\cite{zhao2024neucore}, a model incorporating multi-modal concept alignment and progressive fusion. Baldrati et al.~\cite{baldrati2023zero} explored textual inversion for zero-shot CIR, eliminating the need for task-specific training data. A comprehensive survey by Song et al.~\cite{song2024survey} details CIR models, datasets, and evaluation paradigms.

Despite CIR's advancements, \textit{composed pose retrieval} (CPR) presents unique challenges. Unlike CIR, which modifies object attributes, backgrounds, or styles, our task focuses solely on \textit{human pose}, demanding fine-grained differentiation of body joint movements. Prior work, such as Kim et al.~\cite{kim2021fixmypose}, explored pose retrieval but did not isolate pose-specific modifications, as non-pose attributes were concurrently considered. This makes our task significantly more complex, as body movements are continuous and lack strong texture cues commonly used in CIR. While CIR benefits from shape and background context, our model must detect subtle joint displacements under static scene conditions. Additionally, CIR enjoys large-scale labeled datasets such as FashionIQ~\cite{wu2021fashion} and CIRR~\cite{liu2021cirr}, whereas pose retrieval lacks extensive annotated training data, necessitating automated annotation strategies.\smallskip

\noindent{\bf Harnessing MLLMs for Vision Tasks.} The integration of multimodal large language models (MLLMs) has revolutionized vision tasks by enabling joint visual-textual processing, enhancing image captioning, visual question answering, and cross-modal retrieval. CLIP’s contrastive pretraining facilitates zero-shot classification and retrieval \cite{radford2021learning}, while VisionLLM~v2 unifies vision-language tasks within a single framework \cite{wu2024visionllm}. Techniques such as Visual Context Compression improve efficiency by reducing visual token redundancy \cite{chen2024efficient}, and models like MotionGPT bridge language and motion representation by generating human motion from text \cite{zhang2023motiongpt}.  

Beyond vision, LLMs are leveraged for automated data annotation, alleviating the cost of manual labeling. Studies highlight their effectiveness in generating high-quality annotations \cite{tan2024large}, with selective annotation improving few-shot learning efficiency \cite{su2023selective}. LLM-driven strategies also enhance event extraction by mitigating data scarcity \cite{chen2024large}.  

Building on these advancements, we propose an MLLM-based approach to automate pose transition annotation, integrating visual-textual processing to improve scalability and consistency. By reducing reliance on manual labeling, our method enhances the efficiency and accuracy of pose retrieval systems while addressing subjectivity and annotation bottlenecks.
\section{Method}
\label{sec:method}

\begin{figure*}
\centering
\includegraphics[width=0.9\linewidth]{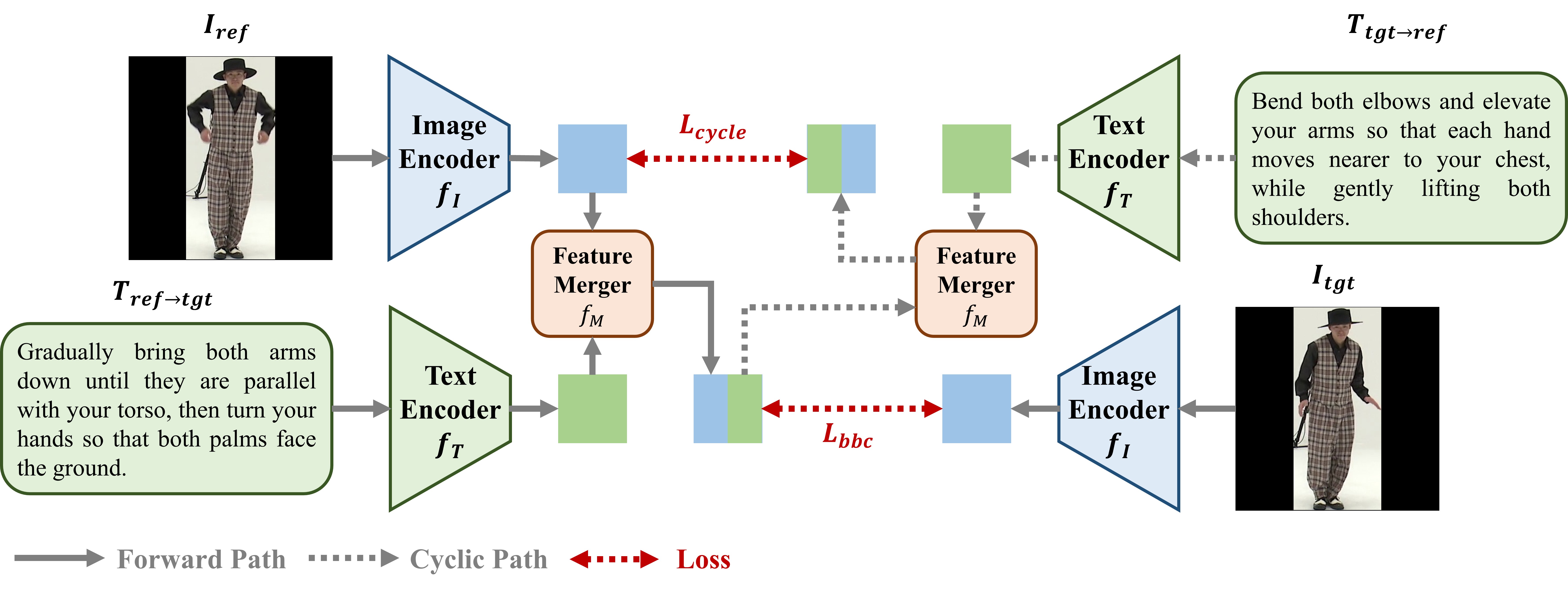}
\caption{During retrieval model training, AutoComPose incorporates a cycle constraint by ensuring that the composed features of input reference images ($I_{ref}$) and transition descriptions ($T_{ref \rightarrow tgt}$) can be transformed back to the reference image features along a cyclic path using the reverse transition descriptions ($T_{tgt \rightarrow ref}$), which are obtained during the generation step.}
\label{fig:cyclic}
\end{figure*}

AutoComPose consists of two primary steps: the \textit{auto}-generation of pose transition descriptions followed by the training of our cyclic model. Each step is elaborated below.

\subsection{Generating Pose Transition Descriptions}
\label{sec:method:genposedes}

The first step of AutoComPose focuses on generating diverse and expressive pose transition descriptions to enhance CPR effectiveness. An overview of this step is presented in Fig.~\ref{fig:AutoComPose}. This process consists of three stages: \smallskip

\noindent{\bf Stage \uppercase\expandafter{\romannumeral 1}: Body Part-based Descriptions.} The effectiveness of pose transition descriptions depends on their ability to capture fine-grained body movements with precision. To achieve this, \textbf{AutoComPose} leverages an \textbf{MLLM} to analyze and compare corresponding body parts between two poses, focusing on key anatomical landmarks: \textit{head, neck, shoulders, arms, elbows, wrists, hands, torso, hips, legs, knees, ankles, and feet}. For each body part undergoing adjustment, the MLLM generates a concise, targeted description of its motion, ensuring clarity and completeness. The selection of these body parts is grounded in their fundamental role in human pose estimation, as established by state-of-the-art benchmarks in pose tracking~\cite{cao2017realtime}, dense pose estimation~\cite{guler2018densepose}, and pose-conditioned motion synthesis~\cite{petrovich2021action}. By explicitly structuring pose transitions at the body-part level, AutoComPose not only enhances interpretability but also offers greater adaptability, making it particularly well-suited for complex motion understanding tasks such as CPR. 

This structured approach mitigates common limitations in holistic descriptions, where subtle yet critical articulations—such as wrist rotation or knee flexion—are frequently overlooked. By maintaining fine-grained motion analysis, AutoComPose provides a richer, more accurate representation of pose transitions, bridging the gap between human perception and computational models. For a detailed breakdown of the MLLM prompting strategy and an evaluation of Stage \uppercase\expandafter{\romannumeral 1}'s effectiveness, we refer the reader to the Supplementary Materials.\smallskip

\noindent{\bf Stage \uppercase\expandafter{\romannumeral 2}: Integration and Diversification.} While body part-based transition descriptions effectively capture fine-grained movement details, they do not naturally align with how users articulate queries. Rather than providing structured, bullet-point descriptions, users typically favor a single, coherent narrative that seamlessly conveys all relevant information. To bridge this gap, AutoComPose prompts the MLLM to synthesize a fluid and concise description that preserves the key motion semantics. Moreover, to account for the inherent variability in human annotations, AutoComPose generates multiple alternative descriptions from the same body part-based inputs, capturing diverse linguistic expressions of the same transition. Additionally, by encouraging the use of analogies when appropriate, AutoComPose enhances expressiveness, making pose transitions more intuitive and relatable. Further details on the prompting strategy can be found in the Supplementary Materials.\smallskip

\noindent{\bf Stage \uppercase\expandafter{\romannumeral 3}: Swapping and Mirroring.} Data augmentation is a standard practice in training image-based machine learning models, with techniques like random flipping~\cite{shorten2019survey} commonly employed to improve robustness. However, in the case of CPR, applying such augmentations is non-trivial, as the corresponding transition descriptions must precisely align with the transformations applied to the images—an effort that traditionally demands additional manual annotations. AutoComPose addresses this challenge by automatically generating consistent transition descriptions through direct manipulation of input image pairs. By default, it applies swapping (time reversal), mirroring (left-right flipping), or both, effectively increasing dataset diversity without requiring extra annotation efforts. By automating these augmentations, AutoComPose not only enhances scalability but also ensures diverse and balanced datasets, making it a practical and efficient solution for CPR.

\subsection{Cyclic CPR Model Training}
\label{sec:method:cyccir}

MLLMs occasionally generate unrelated or incorrect responses, which could degrade retrieval performance when these descriptions are used for training. To mitigate this issue and enhance the robustness of pose retrieval, AutoComPose incorporates a cyclic model training approach, as shown in Fig.~\ref{fig:cyclic}. \smallskip

\noindent{\bf Generic VLM Training.} Assume that we have a reference image $I_{ref}$, a target image $I_{tgt}$, and the corresponding pose transition description $T_{ref \rightarrow tgt}$ generated by AutoComPose. We can fine-tune any off-the-shelf VLMs by projecting $I_{ref}$ and $T_{ref \rightarrow tgt}$ onto a joint manifold using the image encoder $f_{I}$ and text encoder $f_{T}$ of the VLMs. The extracted features are then merged using a feature merger $f_{M}$, which could be as simple as an element-wise sum. The similarity between the composed features and the target image features is then measured using specific metric functions, and this similarity score is used as the training loss. 

In this work, we follow \cite{baldrati2023composed,lee2021cosmo,vo2019composing} and adopt a batch-based classification loss, which is formulated as:
\begin{equation}
    L_{bbc} = \frac{1}{B}\sum_{i=1}^{B}-\log\frac{\exp\{\lambda*\kappa(\psi^i,\phi^i)\}}{\sum_{j=1}^{B}\exp\{\lambda*\kappa(\psi^i,\phi^j)\}},
\end{equation}
where $\psi^i=f_M(f_I(I_{ref}^i),f_T(T_{ref \rightarrow tgt}^i))$ represents the composed features, and $\phi^i=f_I(I_{tgt}^i)$ represents the target image features. Following \cite{baldrati2023composed}, $\kappa$ is defined as the cosine similarity measure, and the parameter $\lambda$ is set to 100. \smallskip

\noindent{\bf Cycle Constraint.} MLLM-generated pose transition descriptions can occasionally introduce errors, such as misidentifying body parts, hallucinating non-existent movements, or omitting crucial motions. As a result, the text features $f_T(T_{ref \rightarrow tgt})$ and $f_T(T_{tgt \rightarrow ref})$ may fail to capture the true correlations between the reference and target image features. This can lead to unreliable composed features and suboptimal model training. One possible solution is to use a more powerful MLLM or fine-tune it on domain-specific data~\cite{hu2022lora}, but this is often resource-intensive or impractical. Another alternative is detecting and correcting incorrect transition descriptions, but this is non-trivial. 

AutoComPose takes a different approach based on a simple hypothesis: if the composed features are correct, they should be reversible using the reverse transition description features. Cycle constraints have been widely used in various vision tasks~\cite{zhu2017unpaired,hoffman2018cycada}, but to our knowledge, they have not been explored in CPR due to the lack of conjugated transition description data (i.e., both forward and backward pose transitions). However, as described in the previous section, AutoComPose can automatically generate the reverse pose transition description $T_{tgt \rightarrow ref}$ by swapping the reference and target images. 

Similar to $L_{bbc}$, AutoComPose defines the cycle constraint loss $L_{cycle}$ as:
\begin{equation}
    L_{cycle} = \frac{1}{B}\sum_{i=1}^{B}-\log\frac{\exp\{\lambda*\kappa(\hat{\psi^i},\hat{\phi^i)}\}}{\sum_{j=1}^{B}\exp\{\lambda*\kappa(\hat{\psi^i},\hat{\phi^j)}\}},
\end{equation}
where $\hat{\psi^i}=f_M(\psi^i,f_T(T_{tgt \rightarrow ref}^i))$ represents the conjugated composed features, and $\hat{\phi^i}=f_I(I_{ref}^i)$ represents the reference image features. The total training loss is then defined as:
\(L_{total} = \omega * L_{bbc} + (1 - \omega) * L_{cycle},\) where $\omega$ is set to 0.5 in this paper.
\section{Experiments}
\label{sec:exp}

\subsection{Datasets and Experimental Setup}
\label{sec:exp:datasetup}

We evaluate AutoComPose using three datasets that encompass a diverse range of body movements:\smallskip

\noindent{\bf FIXMYPOSE.} FIXMYPOSE~\cite{kim2021fixmypose} provides synthetic pose image pairs along with corresponding human-annotated pose transition descriptions, specifically designed for pose correctional captioning and retrieval tasks. The original dataset includes environment-related descriptions (e.g., facing the \emph{stereo system}), which fall outside the scope of this work. To focus specifically on human body poses, we filtered out environment-related descriptions using a large language model, resulting in 1997 training pairs and 510 validation pairs. Since FIXMYPOSE does not publicly release its test set, we used the validation set for our experiments. Please refer to the Supplementary Materials for more details on the filtering process.\smallskip

\noindent{\bf PoseFixCPR.} PoseFix~\cite{delmas2023posefix} provides human-annotated pose transition descriptions for 3D human meshes sampled from AMASS~\cite{mahmood2019amass}, a large-scale dataset of 3D human motions in SMPL format~\cite{SMPL:2015}. Additionally, the dataset includes automatically generated annotations by a rule-based pipeline~\cite{delmas2022posescript} that relies low-level properties of 3D body keypoints. Since the original dataset lacks corresponding 2D images, we followed the BEDLAM~\cite{Black_CVPR_2023} rendering pipeline to synthesize 2D images from SMPL-X~\cite{SMPL-X:2019} bodies using Unreal Engine~\cite{unreal}. To ensure diversity, we rendered images by randomly sampling the following aspects: 111 body shapes~\cite{Patel:CVPR:2021}, 50 skin tones per gender~\cite{meshcapade}, 1738 clothing textures~\cite{Black_CVPR_2023}, 115 HDRI background images~\cite{hdri}, and four different viewing angles (east, north, south, west). This process resulted in 4596 training pairs and 1188 testing pairs, forming a newly generated dataset tailored for the CPR task, which we refer to as \textit{PoseFixCPR}. \smallskip

\noindent{\bf AIST-CPR.} AIST++~\cite{li2021ai} is a large-scale 3D human dance motion dataset containing a wide range of 3D motion sequences along with music. The dataset comprises 1408 video sequences, capturing 30 different dancers performing dances across different genres, music styles, and choreographies. To ensure reasonable and sufficient gaps between paired poses, we constructed pose pairs by first selecting every 15th frames (0.25 seconds apart). Next, we applied a pose estimator~\cite{xu2022vitpose} to filter out pose pairs whose pose distance~\cite{shen2024diversifying} falls outside the predefined range ($[0.1,2.0]$ in this work). This process resulted in 12151 training pairs and 3788 testing pairs, resulting in a new CPR benchmark, namely \textit{AIST-CPR}. Since the dataset does not include transition descriptions for evaluation, we randomly sampled 400 pose pairs from the test set for manual annotation. Each sampled pose pair was annotated by two different  annotators (from a pool of four) to ensure sufficient diversity.\smallskip

\noindent{\bf Generating Descriptions of Pose Transitions.} We leveraged AutoComPose to generate pose transition descriptions for all training pairs across the three datasets mentioned above. For all experiments, we used GPT-4o~\cite{hurst2024gpt} (\textit{gpt-4o-2024-08-06}) as the MLLM, as it was one of the top-performing MLLMs at the time of this study. We also conducted a study using GPT-4o mini to demonstrate AutoComPose's applicability to more accessible MLLMs; results from this study are presented in the Supplementary Materials. By default, AutoCompose generates three descriptions for each pose pair. When swapping and mirroring transformations are applied, this results in a total of 12 descriptions associated with one single pose pair. In rare cases, the MLLM may generate responses that deviate from our guidelines or may even refuse to generate a response (fewer than 2.5\% of cases). We ignored these pairs for simplicity, ensuring that all training pairs used in the model training contain three transition descriptions for each transformation (original, swapping, mirroring, and both).\smallskip

\noindent{\bf Model Architecture and Training Details.} All experiments were conducted based on the implementation from \cite{baldrati2023composed}, using a single NVIDIA GeForce RTX 2080 Ti GPU. We evaluated AutoComPose on CLIP~\cite{radford2021learning} with four different backbones (RN50, RN101, ViT-B/32, and ViT-B/16) to assess its applicability across different computational budgets. During training, we initialized from the official pre-trained models and only fine-tuned the text encoder to reduce the risk of overfitting the image encoder. When multiple descriptions were associated with a pose pair (i.e., due to paraphrasing), we randomly sampled one for each training iteration. As a baseline, we used element-wise sum as the feature merger. Additionally, we tested AutoComPose with Combiner~\cite{baldrati2023composed}, a dedicated feature merger that has demonstrated superior performance compared to na\"ive element-wise sum. We followed a the training procedure similar to \cite{baldrati2023composed}.
First, we fine-tuned the text encoder for 50 epochs with batch size of 128 and a learning rate of $2\mathrm{e}{-6}$. Then, we trained the Combiner while keeping both the image and text encoders fixed for 100 epochs with batch size of 512 and a learning rate of $2\mathrm{e}{-5}$.\smallskip

\noindent{\bf Evaluation Metrics.} For evaluation, we perform retrieval on the entire test set for each dataset. We evaluate model performance using Recall@$k$ ($k \in {1,5,10,50}$), or R@$k$, a standard metric in image retrieval. Recall@$k$ measures the proportion of queries for which the correct target image appears in the top-$k$ retrieved results. Formally, given a query $q$ with its ground-truth target $t$, R@$k$ is defined as:
\begin{equation}
    \text{Recall@}k = \frac{1}{N} \sum_{i=1}^{N} \mathbb{1} \{ t_i \in \mathcal{R}_i^k \},
\end{equation}
where \( N \) is the total number of queries, \( \mathcal{R}_i^k \) is the set of top-$k$ retrieved images for query \( q_i \), and \( \mathbb{1} \{\cdot\} \) is an indicator function. Higher R@$k$ signifies better retrieval performance, reflecting the model’s ability to rank relevant images near the top.

\subsection{Main Results}
\label{sec:exp:mainresults}

\begin{table*}
\centering
\resizebox{0.8\textwidth}{!}{%
\setlength{\tabcolsep}{7.0pt}
\renewcommand{\arraystretch}{1.2}
\begin{tabular}{ccccccccccc}
\toprule[1.3pt]
\multirow{2}{*}{} & \multirow{2}{*}{Backbone} & \multirow{2}{*}{Pose Transition Descriptions} & \multicolumn{4}{c}{FIXMYPOSE (\textit{Size} = 7106)} & \multicolumn{4}{c}{PoseFixCPR (\textit{Size} = 2376)} \\
\cmidrule(lr){4-7} \cmidrule(lr){8-11} & & & R@1 & R@5 & R@10 & R@50 & R@1 & R@5 & R@10 & R@50 \\ \midrule
\multirow{8}*{\rotatebox{90}{CLIP~\cite{radford2021learning}}} & \multirow{2}{*}{RN50} & Human Annotators & 0.20 & 1.57 & 2.94 & 13.53 & 48.73 & 65.15 & 72.14 & 86.11 \\ 
& & AutoComPose & \cellcolor{pastelblue}{\bf 8.24} & 27.45 & 38.63 & 63.53 & 61.36 & 75.84 & 80.22 & 90.32 \\ \cline{2-11}
& \multirow{2}{*}{RN101} & Human Annotators & 3.92 & 12.94 & 19.61 & 45.29 & 59.09 & 72.73 & 79.21 & 88.72 \\ 
& & AutoComPose & 6.47 & \cellcolor{pastelblue}{\bf 32.35} & 43.33 & 71.18 & 63.22 & 76.77 & 82.91 & 91.41 \\ \cline{2-11}
& \multirow{2}{*}{ViT-B/32} & Human Annotators & 5.10 & 20.39 & 29.41 & 54.31 & 54.29 & 66.25 & 72.14 & 85.02 \\ 
& & AutoComPose & 7.06 & 24.90 & 36.08 & 62.16 & 58.08 & 72.31 & 76.94 & 86.87 \\ \cline{2-11}
& \multirow{2}{*}{ViT-B/16} & Human Annotators & 7.45 & 30.20 & 40.98 & 70.59 & 59.26 & 73.74 & 78.62 & 88.30 \\ 
& & AutoComPose & \cellcolor{softgold}{\bf 9.41} & \cellcolor{softgold}{\bf 32.75} & \cellcolor{pastelblue}{\bf 47.45} & 75.49 & 64.14 & 76.26 & 80.13 & 89.56 \\ \toprule[1.3pt]
\multirow{8}*{\rotatebox{90}{CLIP w/ Combiner~\cite{baldrati2023composed}}} & \multirow{2}{*}{RN50} & Human Annotators & 3.14 & 13.14 & 20.98 & 44.12 & 67.93 & 82.32 & 86.36 & 94.11 \\ 
& & AutoComPose & \cellcolor{softgold}{\bf 9.41} & 31.76 & 43.92 & 75.49 & 81.40 & 92.68 & 94.95 & 98.15 \\ \cline{2-11}
& \multirow{2}{*}{RN101} & Human Annotators & 5.49 & 19.22 & 27.84 & 56.47 & 72.14 & 84.85 & 89.81 & 96.04 \\ 
& & AutoComPose & 7.64 & 31.37 & 44.90 & \cellcolor{pastelblue}{\bf 75.88} & 80.72 & 93.18 & 95.88 & 98.73 \\ \cline{2-11}
& \multirow{2}{*}{ViT-B/32} & Human Annotators & 4.51 & 16.67 & 24.90 & 52.94 & 63.13 & 77.10 & 82.41 & 92.93 \\ 
& & AutoComPose & 7.45 & 28.63 & 44.51 & 74.12 & \cellcolor{pastelblue}{\bf 82.58} & \cellcolor{softgold}{\bf 94.11} & \cellcolor{pastelblue}{\bf 95.88} & \cellcolor{pastelblue}{\bf 98.74} \\ \cline{2-11}
& \multirow{2}{*}{ViT-B/16} & Human Annotators & 7.06 & 25.10 & 38.24 & 68.82 & 67.85 & 82.66 & 86.87 & 94.28 \\ 
& & AutoComPose & \cellcolor{pastelblue}{\bf 8.24} & 31.57 & \cellcolor{softgold}{\bf 47.65} & \cellcolor{softgold}{\bf 81.37} & \cellcolor{softgold}{\bf 84.09} & \cellcolor{pastelblue}{\bf 93.43} & \cellcolor{softgold}{\bf 96.30} & \cellcolor{softgold}{\bf 99.07} \\ \bottomrule[1.3pt]
\end{tabular}
}
\caption{Comparison of pose transition descriptions written by human annotators and those generated by AutoComPose on FIXMYPOSE and PoseFixCPR datasets. The \hlc[softgold]{\bf best} and \hlc[pastelblue]{\bf second-best} results in each column are highlighted.}
\label{tab:main_results}
\end{table*}

\noindent{\bf Comparison with Human Annotators.} 
Table ~\ref{tab:main_results} presents a detailed evaluation comparing AutoComPose-generated annotations to human annotations on the FIXMYPOSE and PoseFixCPR datasets. These results span multiple backbone architectures and different strategies for integrating visual and textual features (i.e., CLIP vs. Combiner). \textit{Notably, across all configurations and datasets, AutoComPose consistently surpasses human annotations, never falling short in any scenario.} This underscores AutoComPose’s ability to generalize effectively, eliminating the need for labor-intensive human annotations. We attribute the higher overall performance on PoseFixCPR to fundamental differences in dataset construction. PoseFixCPR features a substantially smaller gallery (2376 vs. 7106) and more distinct one-to-one pose matches, which simplify retrieval. In contrast, FIXMYPOSE’s larger gallery, combined with lower pose and character diversity, results in many visually similar candidates, increasing retrieval ambiguity.\smallskip


\begin{table}
\centering
\resizebox{0.95\columnwidth}{!}{%
\setlength{\tabcolsep}{6.0pt}
\renewcommand{\arraystretch}{1.2}
\begin{tabular}{ccccccc}
\toprule[1.3pt]
\multirow{2}{*}{} & \multirow{2}{*}{Backbone} & \multirow{2}{*}{\begin{tabular}{c} Pose Transition \\ Descriptions \end{tabular}} & \multicolumn{4}{c}{PoseFixCPR (Size = 2376)} \\
\cmidrule(lr){4-7}  & & & R@1 & R@5 & R@10 & R@50 \\ \midrule
\multirow{8}*{\rotatebox{90}{CLIP~\cite{radford2021learning}}} & \multirow{2}{*}{RN50} & Rule-based & 42.00 & 59.51 & 67.09 & 84.34 \\ 
& & AutoComPose & 61.36 & 75.84 & 80.22 & 90.32 \\ \cline{2-7}
& \multirow{2}{*}{RN101} & Rule-based & 48.32 & 64.39 & 69.95 & 82.92 \\ 
& & AutoComPose & 63.22 & 76.77 & 82.91 & 91.41 \\ \cline{2-7}
& \multirow{2}{*}{ViT-B/32} & Rule-based & 53.45 & 67.34 & 73.48 & 85.44 \\ 
& & AutoComPose & 58.08 & 72.31 & 76.94 & 86.87 \\ \cline{2-7}
& \multirow{2}{*}{ViT-B/16} & Rule-based & 58.00 & 72.98 & 78.11 & 88.30 \\ 
& & AutoComPose & 64.14 & 76.26 & 80.13 & 89.56 \\ \toprule[1.3pt]
\multirow{8}*{\rotatebox{90}{CLIP w/ Combiner~\cite{baldrati2023composed}}} & \multirow{2}{*}{RN50} & Rule-based & 73.15 & 86.62 & 90.07 & 96.46 \\ 
& & AutoComPose & 81.40 & 92.68 & 94.95 & 98.15 \\ \cline{2-7}
& \multirow{2}{*}{RN101} & Rule-based & 71.55 & 86.03 & 89.90 & 95.96 \\ 
& & AutoComPose & 80.72 & 93.18 & \cellcolor{pastelblue}{\bf 95.88} & 98.73 \\ \cline{2-7}
& \multirow{2}{*}{ViT-B/32} & Rule-based & 71.04 & 86.70 & 90.40 & 96.63 \\ 
& & AutoComPose & \cellcolor{pastelblue}{\bf 82.58} & \cellcolor{softgold}{\bf 94.11} & \cellcolor{pastelblue}{\bf 95.88} & \cellcolor{pastelblue}{\bf 98.74} \\ \cline{2-7}
& \multirow{2}{*}{ViT-B/16} & Rule-based & 73.15 & 87.71 & 91.41 & 96.80 \\ 
& & AutoComPose & \cellcolor{softgold}{\bf 84.09} & \cellcolor{pastelblue}{\bf 93.43} & \cellcolor{softgold}{\bf 96.30} & \cellcolor{softgold}{\bf 99.07} \\ \bottomrule[1.3pt]
\end{tabular}
}
\caption{\textbf{Rule-based vs. AutoComPose} on PoseFixCPR, highlighting the advantages of AutoComPose in providing less constrained and generalizable pose transition descriptions.}
\label{tab:comp_auto}
\end{table}

\noindent{\bf Comparison with Rule-based Generation.} Table~\ref{tab:comp_auto} clearly demonstrates the consistent superiority of our AutoComPose over the rule-based pose transition descriptions of PoseFix \cite{delmas2023posefix}, across various backbone architectures and combiners. While PoseFix aimed to automate pose description generation, its approach ultimately relied on predefined aggregation rules and template-based sentences, making it inherently rule-based. Furthermore, PoseFix depended on a constrained set of ‘paircodes’—descriptors anchored to absolute 3D keypoint positions in image pairs—which fundamentally restricted the expressiveness and generalizability of the generated descriptions. This rigid formulation limited its adaptability to diverse poses and transitions. In contrast, AutoComPose dynamically constructs descriptions without such constraints, enabling more flexible and contextually rich representations. The substantial performance gap observed in Table 2 further underscores the importance of allowing greater descriptive freedom, reinforcing the effectiveness of our approach.\smallskip

\begin{table}
\centering
\resizebox{0.95\columnwidth}{!}{%
\setlength{\tabcolsep}{6.0pt}
\renewcommand{\arraystretch}{1.3}
\begin{tabular}{ccccccc}
\toprule[1.3pt]
\multirow{2}{*}{} & \multirow{2}{*}{Backbone} & \multirow{2}{*}{\begin{tabular}{c} Pose Transition \\ Descriptions \end{tabular}} & \multicolumn{4}{c}{AIST-CPR (Size = 8409)} \\
\cmidrule(lr){4-7}  & & & R@1 & R@5 & R@10 & R@50 \\ \midrule
\multirow{4}*{\rotatebox{90}{CLIP~\cite{radford2021learning}}} & RN50 & \multirow{4}*{AutoComPose} & 3.25 & 13.38 & 22.62 & 50.75 \\ 
& RN101 & & 5.12 & 16.25 & 23.00 & 52.88 \\ 
& ViT-B/32 & & 3.13 & 9.62 & 16.00 & 42.88 \\
& ViT-B/16 & & 3.38 & 12.50 & 22.00 & 49.13 \\ \toprule[1.3pt]
\multirow{4}*{\rotatebox{90}{Combiner~\cite{baldrati2023composed}}} & RN50 & \multirow{4}*{AutoComPose} & 5.75 & 20.75 & 31.25 & 65.87 \\ 
& RN101 & & \cellcolor{softgold}{\bf 7.00} & \cellcolor{softgold}{\bf 22.37} & \cellcolor{pastelblue}{\bf 32.62} & \cellcolor{pastelblue}{\bf 68.12} \\ 
& ViT-B/32 & & 5.75 & 17.87 & 28.50 & 63.63 \\ 
& ViT-B/16 & & \cellcolor{pastelblue}{\bf 6.75} & \cellcolor{pastelblue}{\bf 21.13} & \cellcolor{softgold}{\bf 32.75} & \cellcolor{softgold}{\bf 69.63} \\ \bottomrule[1.3pt]
\end{tabular}
}
\caption{Evaluation on AIST-CPR using descriptions generated via AutoComPose.}
\label{tab:main_aist}
\end{table}

\noindent{\bf Results on AIST-CPR.}
We present retrieval performance results across a comprehensive set of feature merger (CLIP vs. Combiner) combined with various backbone architectures on the AIST-CPR. Notably, the AIST-CPR dataset lacks human-annotated transition description for the train set, highlighting the importance of our methodological approach in generating and utilizing such annotations.

\subsection{Additional Analyses}
\label{sec:exp:analyzes}

\begin{table}
\centering
\resizebox{0.95\columnwidth}{!}{%
\setlength{\tabcolsep}{6.0pt}
\renewcommand{\arraystretch}{1.2}
\begin{tabular}{ccccccc}
\toprule[1.3pt]
& Dataset & AutoComPose & R@1 & R@5 & R@10 & R@50 \\ \midrule
\multirow{9}*{\rotatebox{90}{CLIP (RN50)}} & \multirow{3}{*}{FIXMYPOSE} & Our (full) & \cellcolor{softgold}{\bf 8.24} & \cellcolor{softgold}{\bf 27.45} & \cellcolor{softgold}{\bf 38.63} & \cellcolor{softgold}{\bf 63.53} \\
& & \textcolor{red}{\bf (-)} Cyclic & 5.88 & 20.00 & 32.35 & 55.10 \\
& & \textcolor{red}{\bf (-)} SW \& MI & 2.35 & 9.02 & 14.31 & 34.51 \\ \cline{2-7}
& \multirow{3}{*}{PoseFixCPR} & Our (full) & \cellcolor{softgold}{\bf 61.36} & \cellcolor{softgold}{\bf 75.84} & \cellcolor{softgold}{\bf 80.22} & \cellcolor{softgold}{\bf 90.32} \\
& & \textcolor{red}{\bf (-)} Cyclic & 56.48 & 74.07 & 78.20 & 89.56 \\
& & \textcolor{red}{\bf (-)} SW \& MI & 48.15 & 64.65 & 73.06 & 86.45\\ \cline{2-7}
& \multirow{3}{*}{AIST-CPR} & Our (full) & \cellcolor{softgold}{\bf 3.25} & \cellcolor{softgold}{\bf 13.38} & \cellcolor{softgold}{\bf 22.62} & \cellcolor{softgold}{\bf 50.75} \\
& & \textcolor{red}{\bf (-)} Cyclic & 3.13 & 12.00 & 19.75 & 44.37 \\
& & \textcolor{red}{\bf (-)} SW \& MI & 3.00 & 7.75 & 14.25 & 36.25 \\ \toprule[1.3pt]
\multirow{9}*{\rotatebox{90}{CLIP (RN101)}} & \multirow{3}{*}{FIXMYPOSE} & Our (full) & \cellcolor{softgold}{\bf 6.47} & \cellcolor{softgold}{\bf 32.35} & \cellcolor{softgold}{\bf 43.33} & \cellcolor{softgold}{\bf 71.18} \\
& & \textcolor{red}{\bf (-)} Cyclic & \cellcolor{softgold}{\bf 6.47} & 24.51 & 39.41 & 64.31 \\
& & \textcolor{red}{\bf (-)} SW \& MI & 5.88 & 18.82 & 30.39 & 56.08 \\ \cline{2-7}
& \multirow{3}{*}{PoseFixCPR} & Our (full) & \cellcolor{softgold}{\bf 63.22} & \cellcolor{softgold}{\bf 76.77} & \cellcolor{softgold}{\bf 82.91} & \cellcolor{softgold}{\bf 91.41} \\
& & \textcolor{red}{\bf (-)} Cyclic & 59.43 & 75.76 & 81.14 & 90.99 \\
& & \textcolor{red}{\bf (-)} SW \& MI & 52.86 & 69.02 & 76.09 & 86.45 \\ \cline{2-7}
& \multirow{3}{*}{AIST-CPR} & Our (full) & 5.12 & \cellcolor{softgold}{\bf 16.25} & 23.00 & \cellcolor{softgold}{\bf 52.88} \\
& & \textcolor{red}{\bf (-)} Cyclic & \cellcolor{softgold}{\bf 5.25} & 15.75 & \cellcolor{softgold}{\bf 23.75} & 49.75 \\
& & \textcolor{red}{\bf (-)} SW \& MI & 3.13 & 11.00 & 16.00 & 39.63 \\ \bottomrule[1.3pt]
\end{tabular}
}
\caption{\textbf{Ablation study} demonstrating the impact of data augmentation and cyclic loss guidance on retrieval accuracy. \textcolor{red}{(-)} symbol denotes the removal of the corresponding component from the architecture. SW and MI refer to swapping and mirroring.}
\label{tab:ablation_components}
\end{table}

\noindent{\bf Ablation Study of Each Component in AutoComPose.} In Table~\ref{tab:ablation_components}, we demonstrate the critical role of dataset diversification and the effectiveness of cyclic loss guidance in optimizing model training. Notably, our results show consistent and substantial improvements in retrieval accuracy across all benchmark datasets and backbone architectures—at times achieving up to a threefold increase (e.g., 9.02 to 27.45). Importantly, \textit{these gains come without any additional computational cost} during inference, making our approach highly efficient and scalable.\smallskip

\begin{table}
\centering
\resizebox{0.9\columnwidth}{!}{%
\setlength{\tabcolsep}{6.0pt}
\renewcommand{\arraystretch}{1.0}
\begin{tabular}{cccccc}
\toprule[1.3pt]
\multirow{2}{*}{Model} & \multirow{2}{*}{Paraphrasing} & \multicolumn{4}{c}{FIXMYPOSE (Size = 7106)} \\
\cmidrule(lr){3-6}  & & R@1 & R@5 & R@10 & R@50 \\ \midrule
\multirow{3}*{CLIP~\cite{radford2021learning}} & 1 & 5.88 & 20.39 & 30.20 & 54.31 \\ 
& 3 (default) & 8.24 & 27.45 & 38.63 & 63.53 \\ 
& 5 & \cellcolor{pastelblue}{\bf 9.02} & 28.63 & 39.02 & 64.71 \\ \toprule[1.3pt]
\multirow{3}*{Combiner~\cite{baldrati2023composed}} & 1 & 7.65 & 27.65 & 40.00 & 67.06 \\ 
& 3 (default) & \cellcolor{softgold}{\bf 9.41} &\cellcolor{softgold}{\bf 31.76} & \cellcolor{softgold}{\bf 43.92} & \cellcolor{pastelblue}{\bf 75.49} \\ 
& 5 & 8.82 & \cellcolor{pastelblue}{\bf 31.37} & \cellcolor{pastelblue}{\bf 43.53} & \cellcolor{softgold}{\bf 77.45} \\ \bottomrule[1.3pt]
\end{tabular}
}
\caption{\textbf{Impact of Paraphrasing.} The numbers 1, 3, and 5 indicate the count of paraphrased descriptions per pose transition.}
\label{tab:n_sents}
\end{table}

\noindent{\bf Impact of Paraphrasing.} We conducted an empirical study to rigorously assess the impact of description diversification via paraphrasing (see Stage II in Section~\ref{sec:method}), where AutoComPose generates diverse linguistic expressions for the same pose transition. As shown in Table~\ref{tab:n_sents}, we observe a consistent performance improvement, confirming that diversification enhances retrieval accuracy. This result supports the hypothesis that accounting for natural language variability improves model robustness, better aligning with the variations in user queries in real-world scenarios.\smallskip

\begin{table}
\centering
\resizebox{\columnwidth}{!}{%
\setlength{\tabcolsep}{6.0pt}
\renewcommand{\arraystretch}{1.0}
\begin{tabular}{cccccc}
\toprule[1.3pt]
\multirow{2}{*}{\begin{tabular}{c} Pose Transition \\ Descriptions \end{tabular}} & \multirow{2}{*}{Paraphrasing} & \multicolumn{4}{c}{PoseFixCPR (Size = 2376)} \\
\cmidrule(lr){3-6} & & R@1 & R@5 & R@10 & R@50 \\ \midrule
\multirow{2}{*}{Human} & 1 & 48.73 & 65.15 & 72.14 & 86.11 \\
 & 3 & 52.02 & \cellcolor{pastelblue}{\bf 71.55} & \cellcolor{pastelblue}{\bf 77.44} & \cellcolor{pastelblue}{\bf 88.55} \\ \midrule
\multirow{3}{*}{Rule-based} & 1 & 31.73 & 50.42 & 58.84 & 77.10 \\
 & 3 & 42.00 & 59.51 & 67.09 & 84.34 \\
 & 3 (\textcolor{red}{w/} Cyclic) & \cellcolor{pastelblue}{\bf 55.05} & 70.20 & 75.67 & 88.47 \\ \midrule
AutoComPose & 3 & \cellcolor{softgold}{\bf 61.36} & \cellcolor{softgold}{\bf 75.84} & \cellcolor{softgold}{\bf 80.22} & \cellcolor{softgold}{\bf 90.32} \\ \bottomrule[1.3pt]
\end{tabular}
}
\caption{{\bf Paraphrasing and Cyclic Training} for human-annotated and rule-based descriptions (following the CLIP-RN50 setting).}
\label{tab:rebuttal_w23_comb_v2}
\end{table}

\noindent{\bf Paraphrasing and Cyclic Training for Other Methods.} We evaluated the impact of paraphrasing and cyclic training on human-annotated and rule-based transition descriptions. Official paraphrases and reverse descriptions (via the pipeline in~\cite{delmas2023posefix}) were used. As shown in Table~\ref{tab:rebuttal_w23_comb_v2}, paraphrasing improved performance in both cases, while AutoComPose remained the most effective due to its superior design. Moreover, cyclic training enhanced the performance of the rule-based method, confirming its broader applicability. \smallskip

\begin{figure}
\centering
\includegraphics[width=\columnwidth]{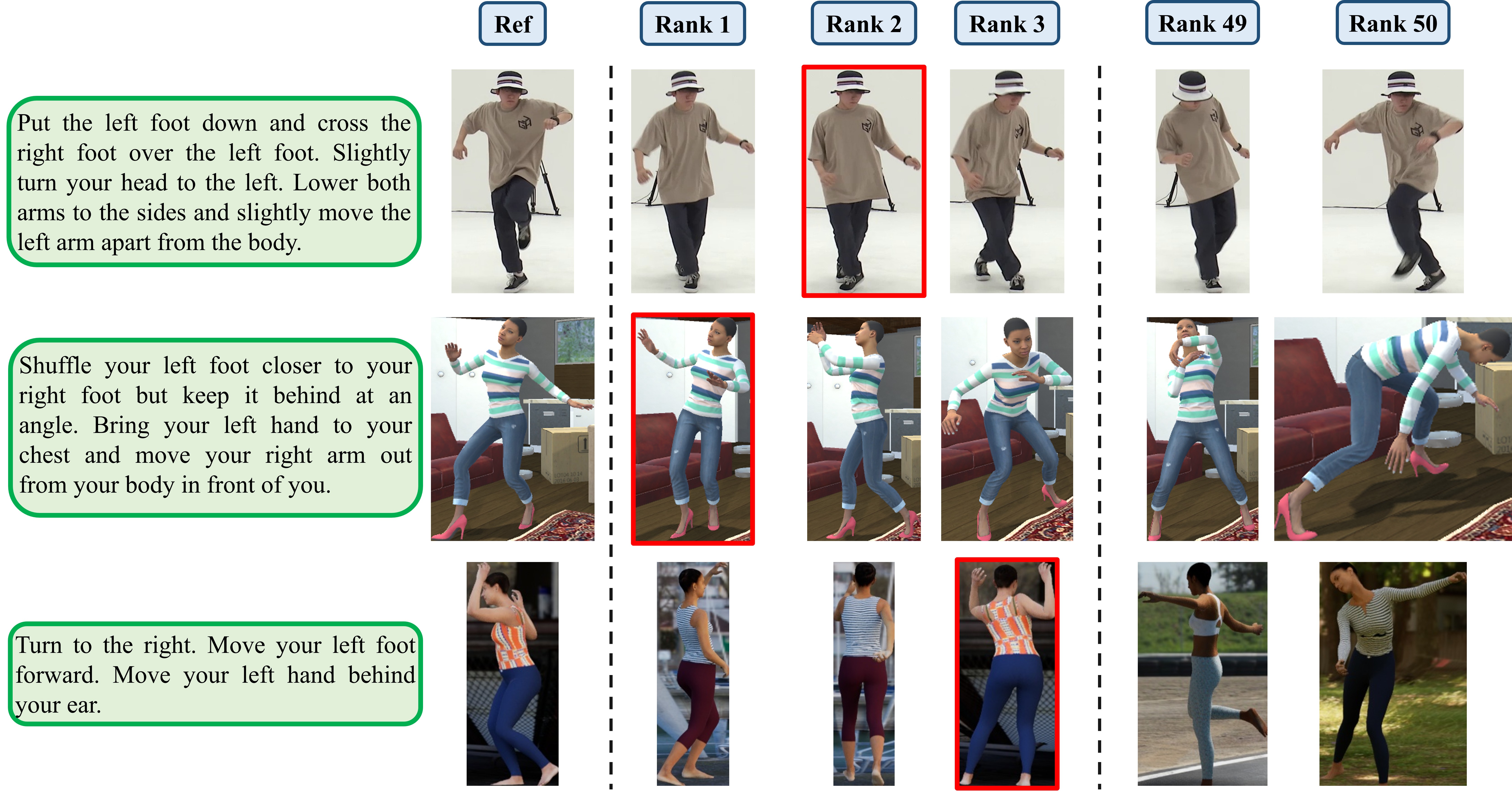}
\caption{\textbf{Example CPR results.} Ranked retrieval results are shown for a given reference image (Ref) and a transitional pose description (green box) as the query. Red boxes indicate the correct target images. The top, middle, and bottom rows correspond to results from AIST-CPR, FIXMYPOSE, and PoseFixCPR.}
\label{fig:qualitative}
\end{figure}

\noindent{\bf Qualitative Results.} Fig.~\ref{fig:qualitative} presents example retrieval results obtained using our model trained on transition descriptions from AutoComPose. While the model is not flawless, we observe that images aligning with the given [Reference image + transition description] are ranked among the top results, whereas less relevant images are positioned further down the ranking. Additional qualitative results are provided in the Supplementary Materials.\smallskip

\noindent{\bf Conclusion and Broader Impact.} \label{sec:conclusion} We introduced \textit{AutoComPose}, the first MLLM-based framework for automatically generating pose transition descriptions, eliminating the need for costly human annotations and enhancing CPR accessibility. AutoComPose is dataset-agnostic, integrating seamlessly with any human image dataset. To foster future research, we release AIST-CPR and PoseFixCPR, as new testbeds for CPR evaluation. Beyond CPR, our work demonstrates MLLMs' potential for structured motion understanding, reducing annotation costs while improving scalability. By introducing benchmarks and a scalable annotation framework, we lay the foundation for advancing retrieval tasks and encourage broader adoption of auto-generated pose descriptions in vision-language research.\smallskip
\noindent{\bf Acknowledgment.} This research was sponsored by the Defense Threat Reduction Agency (DTRA) and the DEVCOM Army Research Laboratory (ARL) under Grant No. W911NF2120076. This research was also sponsored in part by the Army Research Office and Army Research Laboratory (ARL) under Grant Number W911NF-21-1-0258. The views and conclusions contained in this document are those of the authors and should not be interpreted as representing the official policies, either expressed or implied, of the Army Research Office, Army Research Laboratory (ARL) or the U.S. Government. The U.S. Government is authorized to reproduce and distribute reprints for Government purposes notwithstanding any copyright notation herein.

\appendix

\section{Additional Results}
\label{suppl:additional}

\begin{figure}
\centering
\includegraphics[width=\columnwidth]{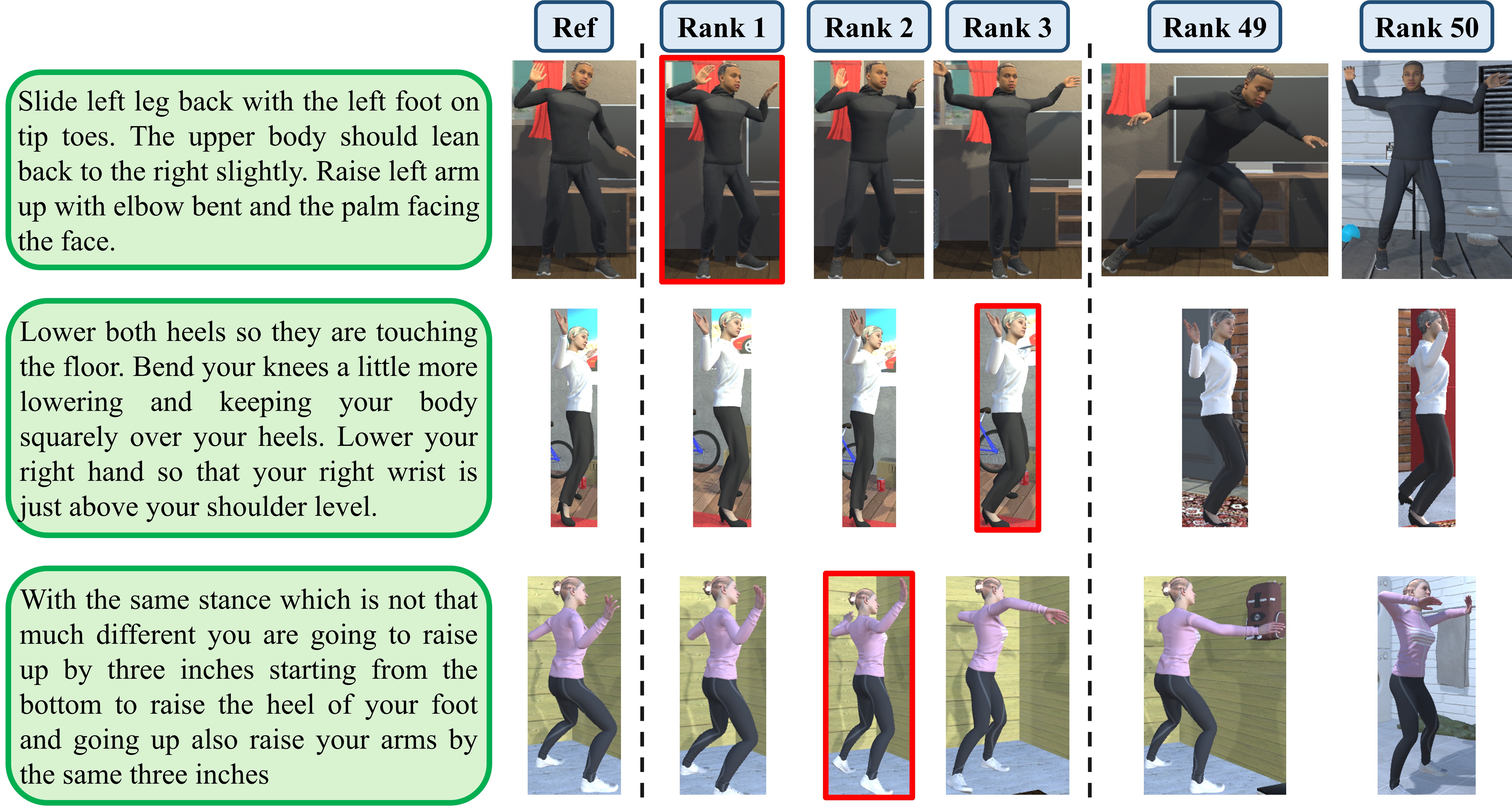}
\caption{\textbf{Additional CPR results on FIXMYPOSE.} Ranked retrieval results are shown for a given reference image (Ref) and a transitional pose description (green box) as the query. Red boxes indicate the correct target images.}
\label{fig:qualitative_fixmypose}
\end{figure}

\begin{figure}
\centering
\includegraphics[width=\columnwidth]{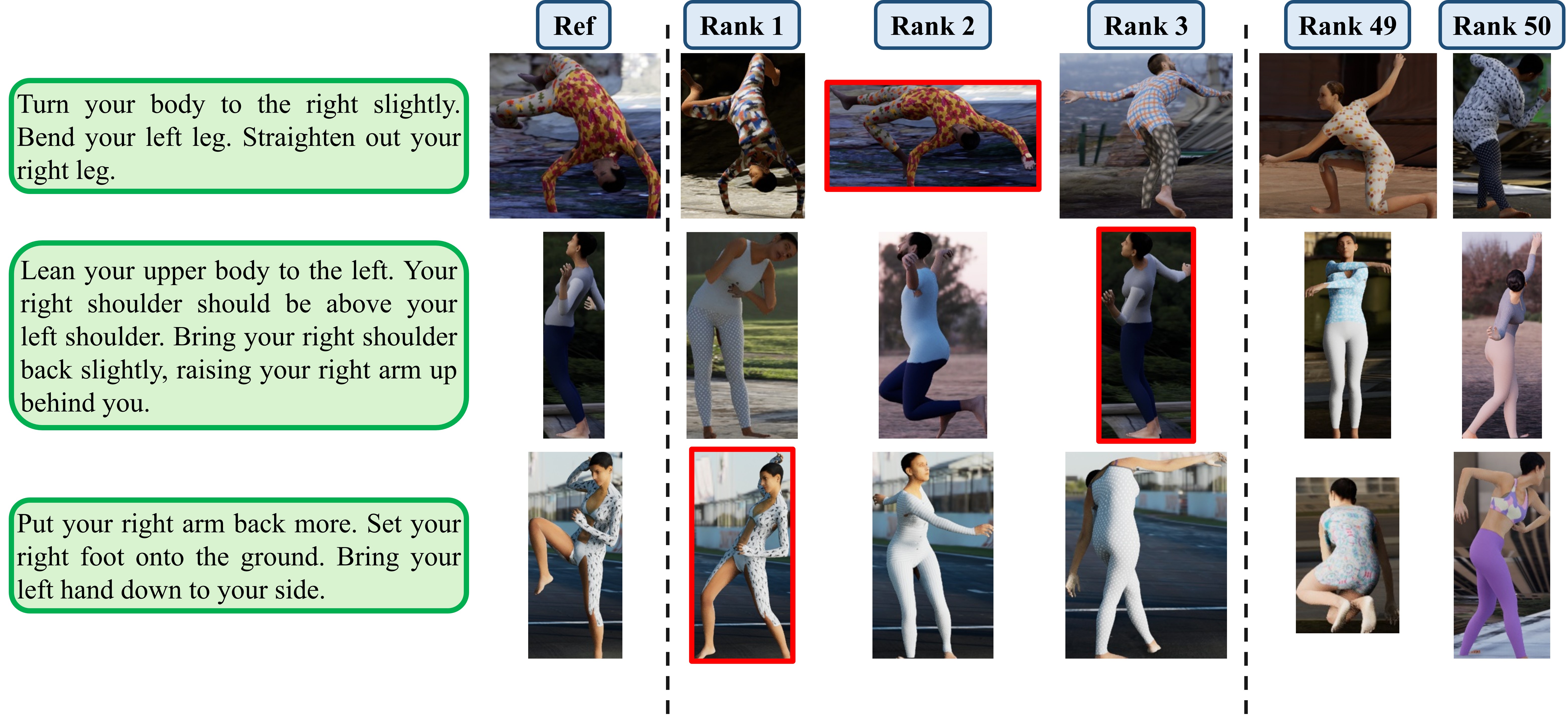}
\caption{\textbf{Additional CPR results on PoseFixCPR.} Ranked retrieval results are shown for a given reference image (Ref) and a transitional pose description (green box) as the query. Red boxes indicate the correct target images.}
\label{fig:qualitative_posefix}
\end{figure}

\begin{figure}
\centering
\includegraphics[width=\columnwidth]{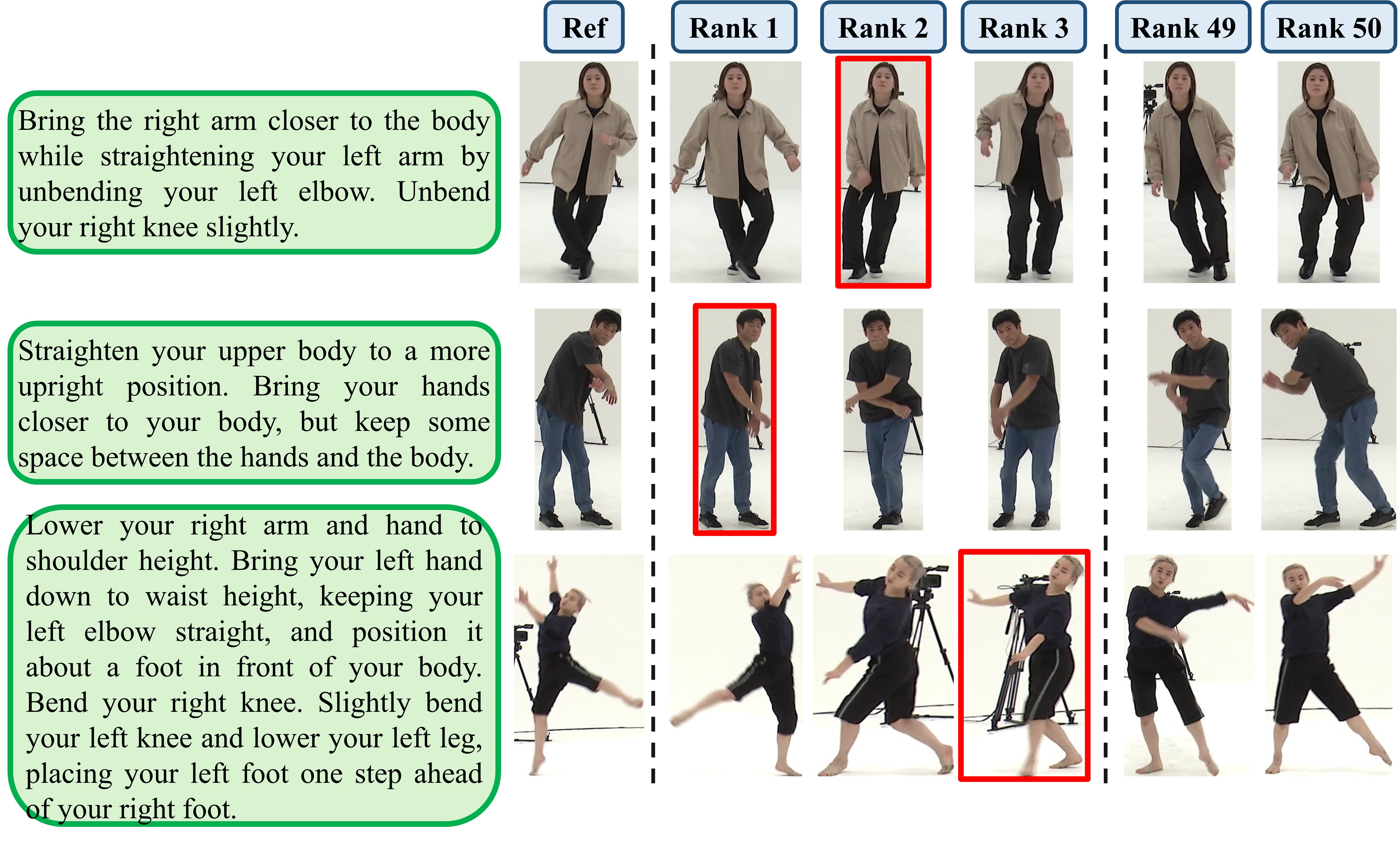}
\caption{\textbf{Additional CPR results on AIST-CPR.} Ranked retrieval results are shown for a given reference image (Ref) and a transitional pose description (green box) as the query. Red boxes indicate the correct target images.}
\label{fig:qualitative_aist}
\end{figure}

\noindent{\bf Qualitative Results.} Similar to Figure~4 in the main manuscript, we present additional qualitative results for FIXMYPOSE (Fig.~\ref{fig:qualitative_fixmypose}), PoseFixCPR (Fig.~\ref{fig:qualitative_posefix}), and AIST-CPR (Fig.~\ref{fig:qualitative_aist}), respectively. The observed trend remains consistent: images that closely align with the provided reference image and transition description are ranked higher, while less relevant images are ranked lower.\smallskip

\noindent{\bf Effectiveness of Stage \uppercase\expandafter{\romannumeral 1}.} To validate the effectiveness of Stage \uppercase\expandafter{\romannumeral 1} in AutoComPose (i.e., body part-based descriptions), we modified the original prompts used by AutoComPose (Figures~\ref{fig:suppl-prompts-parts}~and~\ref{fig:suppl-prompts-intdiv}) to exclude body part-related queries (Figures~\ref{fig:suppl-prompts-no-parts}~and~\ref{fig:suppl-prompts-no-parts-intdiv}). As shown in Table~\ref{tab:rebuttal_w14_comb_v2}, removing Stage \uppercase\expandafter{\romannumeral 1} results in a noticeable drop in performance, underscoring its critical role.\smallskip

\noindent{\bf Smaller, more accessible MLLM.} We conducted an experiment using GPT-4o mini (\textit{gpt-4o-mini-2024-07-18}), a significantly smaller model that is over 16 times more cost-efficient than GPT-4o. As shown in Table~\ref{tab:rebuttal_w14_comb_v2}, the impact of AutoComPose remains substantial, demonstrating its effectiveness and applicability even with lightweight models.

\begin{table}
\centering
\resizebox{0.8\columnwidth}{!}{%
\setlength{\tabcolsep}{6.0pt}
\renewcommand{\arraystretch}{1.0}
\begin{tabular}{ccccc}
\toprule[1.3pt]
\multirow{2}{*}{\begin{tabular}{c} Pose Transition \\ Descriptions \end{tabular}} & \multicolumn{4}{c}{FIXMYPOSE (Size = 7106)} \\
 & R@1 & R@5 & R@10 & R@50 \\ \midrule
AutoComPose (full) & \cellcolor{pastelblue}{\bf 8.24} & \cellcolor{softgold}{\bf 27.45} & \cellcolor{softgold}{\bf 38.63} & \cellcolor{softgold}{\bf 63.53} \\ 
\textcolor{red}{(-)} Stage \uppercase\expandafter{\romannumeral 1} & \cellcolor{softgold}{\bf 8.63} & \cellcolor{pastelblue}{\bf 26.67} & \cellcolor{pastelblue}{\bf 37.84} & \cellcolor{pastelblue}{\bf 62.75} \\
\textcolor{red}{w/} GPT-4o mini & 6.67 & 19.80 & 30.20 & 56.86 \\ \midrule
Human & 0.20 & 1.57 & 2.94 & 13.53 \\ \bottomrule[1.3pt]
\end{tabular}
}
\caption{{\bf Additional Ablation Study on AutoComPose.} The results are reported using the CLIP-RN50 setting.}
\label{tab:rebuttal_w14_comb_v2}
\end{table}

\section{Dataset Details}
\label{suppl:datasets}

\begin{figure*}
\centering
\includegraphics[width=0.95\linewidth]{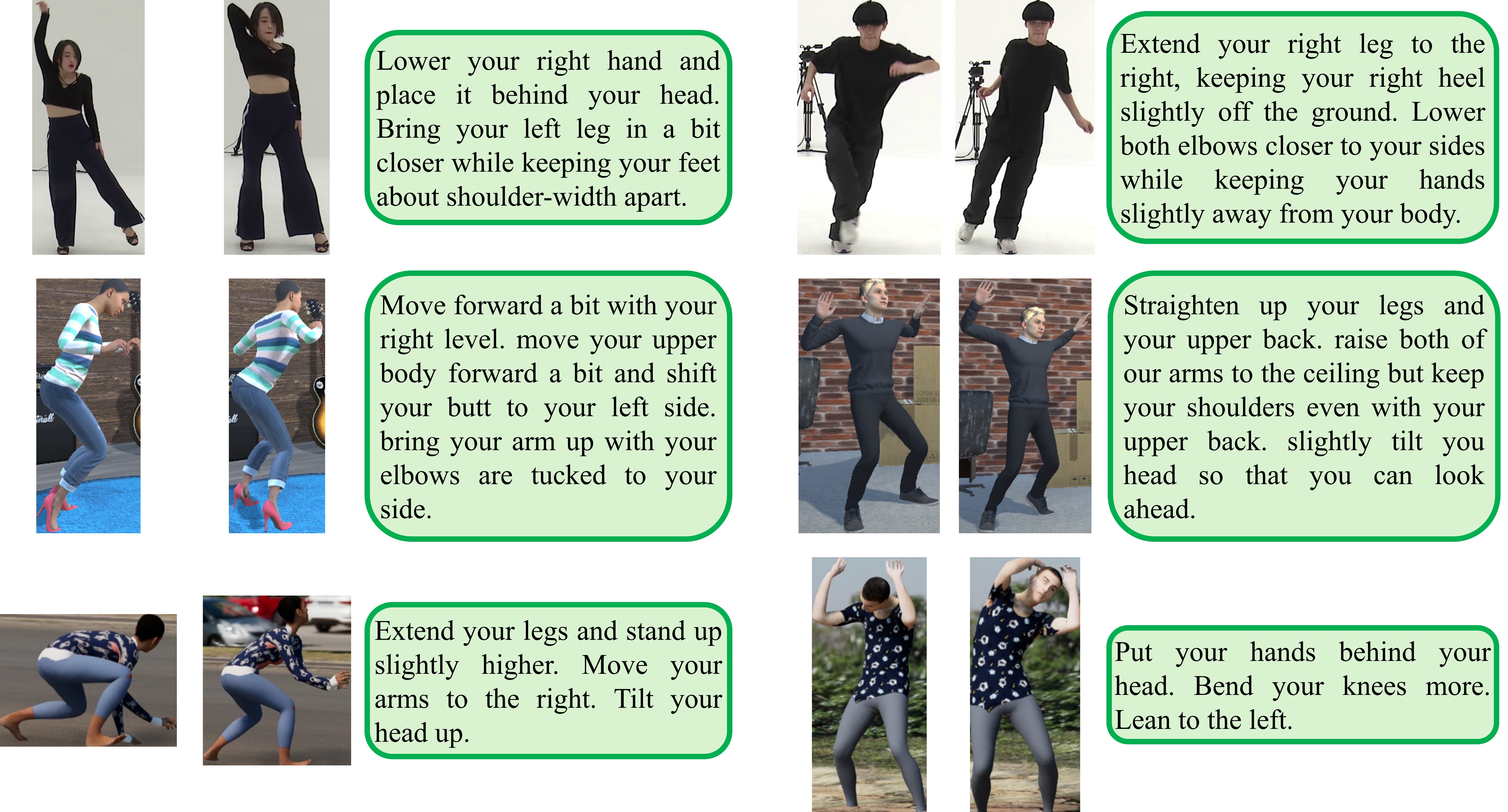}
\caption{{\bf Example pairs from three datasets.} The top, middle, and bottom rows correspond to the pairs from AIST-CPR, FIXMYPOSE, and PoseFixCPR, respectively.}
\label{fig:dataset_all}
\end{figure*}

\noindent{\bf Triplet Examples.} We showcase example pairs from the three datasets in Fig.~\ref{fig:dataset_all}.\smallskip

\begin{figure}
\centering
\includegraphics[width=\columnwidth]{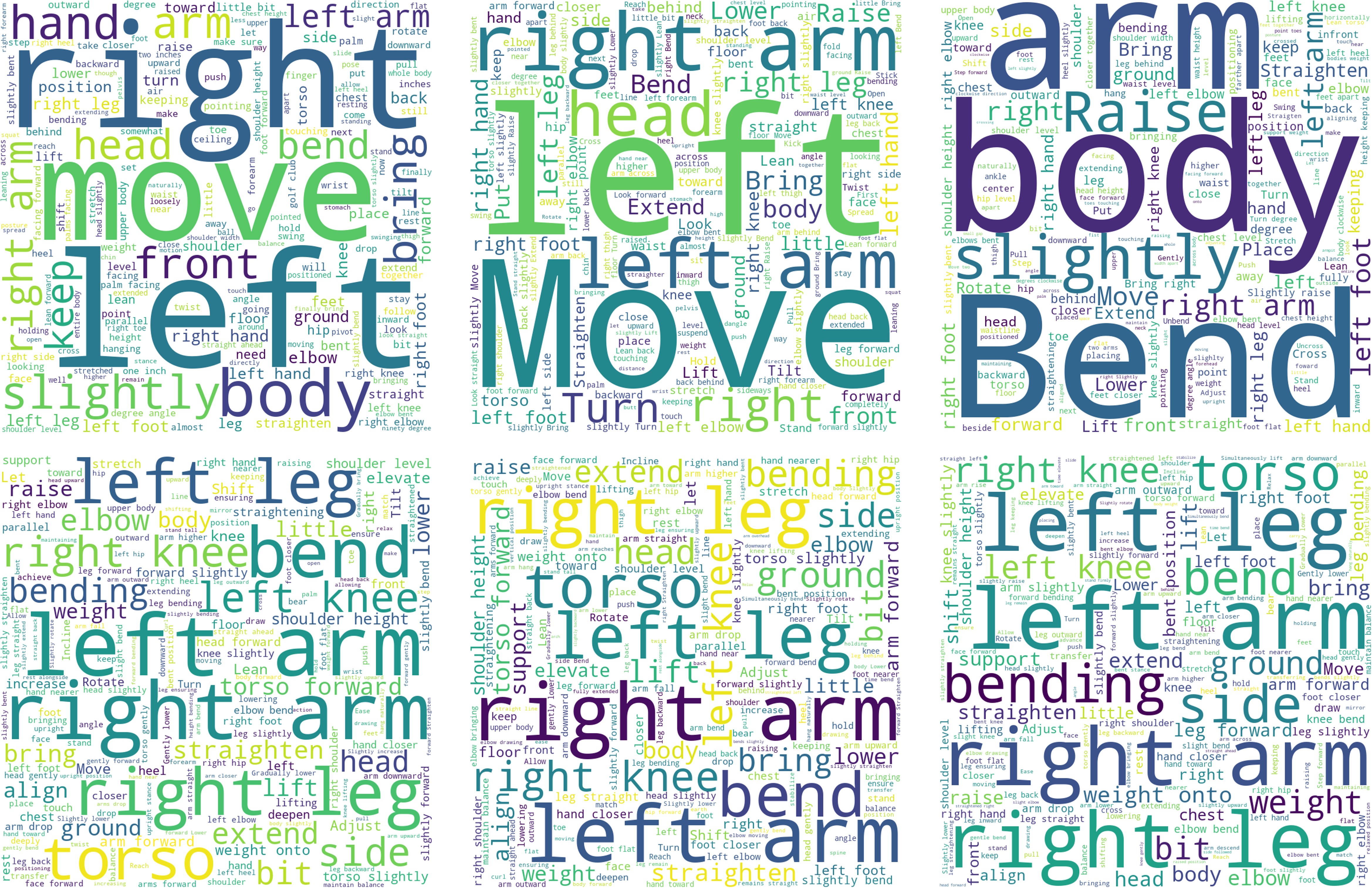}
\caption{{\bf Word clouds} generated from human-annotated texts (top) and AutoComPose-generated texts (bottom) across three datasets: FIXMYPOSE (left), PoseFixCPR (middle), and AIST-CPR (right).}
\label{fig:clouds}
\end{figure}

\noindent{\bf Word Clouds.} We compared word clouds generated from texts produced by AutoComPose with those from human annotators across the three datasets we used, as shown in Fig~\ref{fig:clouds}. Our findings indicate that texts generated by AutoComPose exhibit greater diversity and less bias. \smallskip

\noindent{\bf FIXMYPOSE Filtering.} A portion of the pose transition descriptions provided by FIXMYPOSE~\cite{kim2021fixmypose} contains environment-related instructions, such as guiding a pose transition by referring certain objects in a scene. Since our focus in this paper is on generating pose transition descriptions solely based on body movements, we utilized the multimodal large language model (MLLM)~\cite{hurst2024gpt} with \emph{Prompt-1} (Fig.~\ref{fig:suppl-prompts-envs}) to detect and filter out such descriptions. Three examples of environment-related descriptions were included in the prompt to guide the filtering process. \smallskip

\begin{figure*}
\centering
\begin{tcolorbox}[title=Prompt-1]
You will be given an instruction guiding an individual to transition from their current position to a target position.\\
Your task is to determine whether the instruction includes any environmental direction descriptions—guiding the individual to move relative to a specific object nearby.\\
For example, the following instructions contain environmental direction descriptions, highlighted in quotation marks:
\begin{enumerate}
    \item bring your right foot outwards at a 90 degree angle. bring your left and right hand up like a gun is being pointed at you "keep your position facing the stereo system".
    \item "lean your upper body forward to the plant pot" move the right leg to the right side a little bend the left leg a bit more move the right hand from waist level to beside the hip.
    \item keep your feet and your hands on the floor as support push your arms up move your body and your legs up towards the ceiling "lift up your head a little facing the rug".
\end{enumerate}
In contrast, the following instructions do not contain such descriptions:
\begin{enumerate}
    \item position both your hands in front of your face is if you are grasping a pole. straighten the right leg so that the foot is flat on the floor. bend your left knee a little while pivoting on the ball of your foot. your head should face your hands.
    \item first put your feet together  take your right foot and put it slightly behind you. while keeping your lower body facing forward  twist your upper body to the left. then hold your hands in the air at about the level of your head and turn your head to the left as well.
    \item lift your right foot and pull it behind you. extend both of your hands gently outwards as if you are balancing on a surfboard while it is going under a big wave.
\end{enumerate}
If the instruction contains environmental direction descriptions, respond with "Yes." Otherwise, respond with "No."\\
Provide only the requested answer ("Yes" or "No") with no additional text before or after.
\end{tcolorbox}
\caption{The prompt for removing environment-related descriptions in the FIXMYPOSE dataset.}
\label{fig:suppl-prompts-envs}
\end{figure*}

\begin{figure*}
\centering
\includegraphics[width=\linewidth]{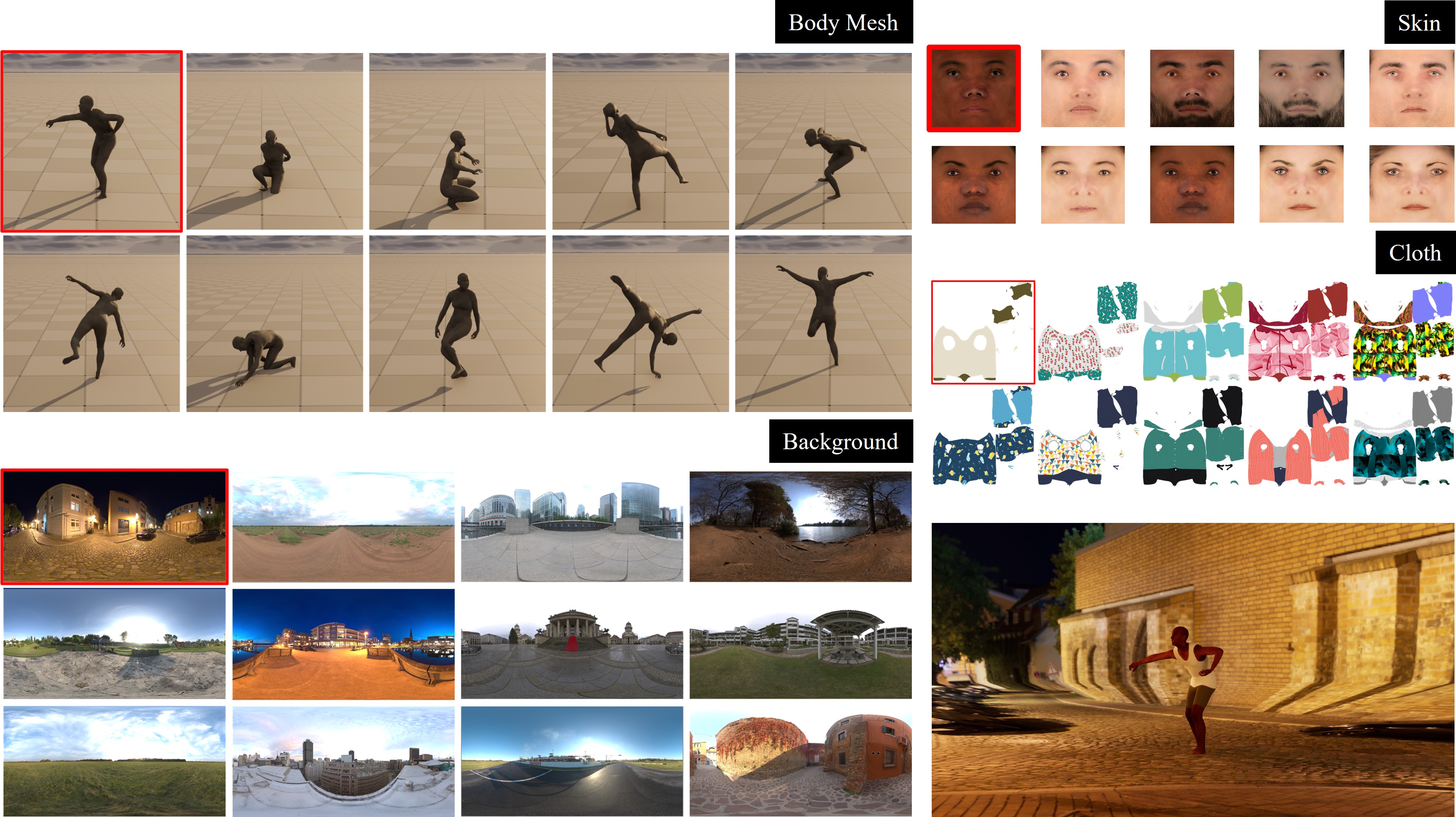}
\caption{{\bf Sampled assets used for constructing PoseFixCPR.} The highlighted assets (in the red box) were used to render the bottom-right image.}
\label{fig:rendering}
\end{figure*}

\begin{figure*}
\centering
\includegraphics[width=\linewidth]{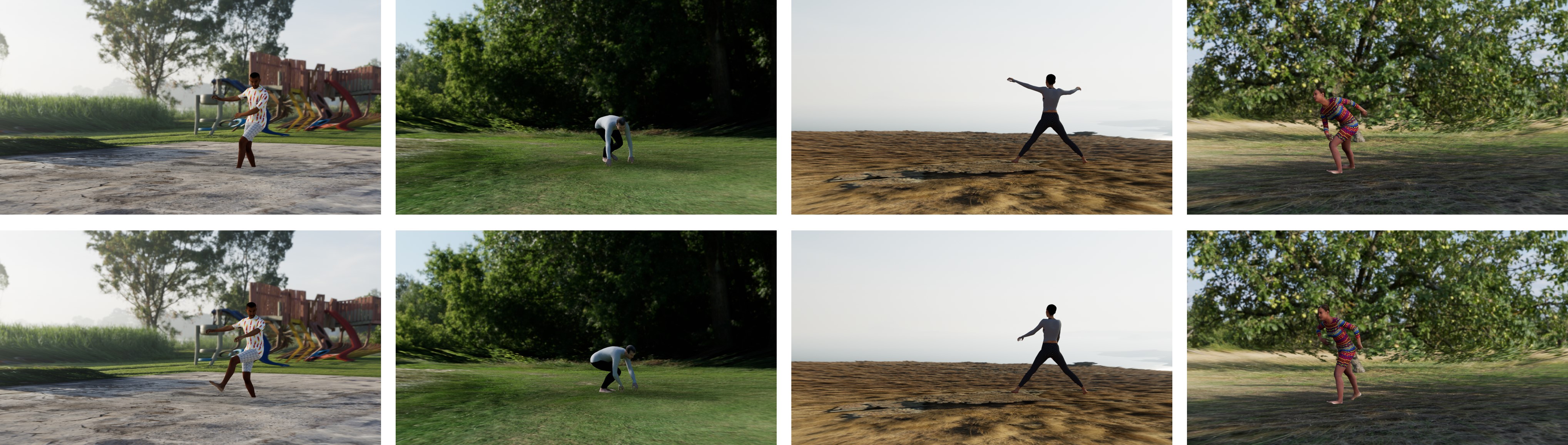}
\caption{{\bf Sampled images from PoseFixCPR.}}
\label{fig:rendering-outputs}
\end{figure*}

\noindent{\bf PoseFixCPR Rendering.} We constructed PoseFixCPR by rendering 2D images (including masks) obtained from 3D pose pairs from PoseFix~\cite{delmas2023posefix} using Unreal Engine~\cite{unreal} based on the BEDLAM~\cite{Black_CVPR_2023} rendering pipeline. A subset of assets used for rendering---including body meshes, body and clothing UV texture maps, and high-dynamic range panoramic images (HDRIs) that used for image-based lighting---is shown in Fig.~\ref{fig:rendering}. A selection of output images is displayed in Fig.~\ref{fig:rendering-outputs}. Note that only the human regions are used for composed pose retrieval (CPR), which can be extracted using the generated masks. \smallskip

\section{Generating Pose Transition Descriptions}
\label{suppl:details_autocompose}

\noindent{\bf MLLM Prompts.} AutoComPose utilizes two prompts to communicate with the multimodal large language model (MLLM), as described in the main manuscript. \emph{Prompt-2} (Fig.~\ref{fig:suppl-prompts-parts}) is designed to guide the MLLM in generating body part-based pose transition descriptions. \emph{Prompt-3} (Fig.~\ref{fig:suppl-prompts-intdiv}) is used to helps the MLLM integrate these body part-based descriptions into more natural and complete sentences. \smallskip

\begin{figure*}
\centering
\begin{tcolorbox}[title=Prompt-2]
You are a skilled fitness\textbackslash dance instructor specializing in guiding individuals through pose transitions. \\
You will receive an image showing two poses side by side: the starting pose on the left and the target pose on the right. \\ 
Your task is to describe the transition from the starting pose to the target pose. \\
To achieve this:
\begin{itemize}[label=-]
\item Compare the following body parts in the two poses: head, neck, left shoulder, right shoulder, left arm, right arm, left elbow, right elbow, left wrist, right wrist, left hand, right hand, torso, left hip, right hip, left leg, right leg, left knee, right knee, left ankle, right ankle, left foot, right foot.
\item For each body part that requires adjustment, provide a clear and concise one-sentence description of the movement.
\item Structure your description as follows: "1. Right Arm: Lift the right arm above the head, transitioning smoothly from its resting position." "2. Left Leg: Straighten the left leg fully as it supports the body’s weight in standing position."
\end{itemize}
Below are some important guidelines that must be followed:
\begin{itemize}[label=-]
\item Mention as few body parts as possible; omit subtle or nearly imperceptible movements.
\item If the person is facing the camera, the left hand of the person is defined as the hand on your (the viewer's) right side.
\item Provide only the requested descriptions of body part transitions, with no additional information before or after.
\end{itemize}
\end{tcolorbox}
\caption{The prompt for generating body part-based pose transition descriptions.}
\label{fig:suppl-prompts-parts}
\end{figure*}

\begin{figure*}
\centering
\begin{tcolorbox}[title=Prompt-3]
You will be given a set of bullet points describing a specific human pose transition. \\
Your task is to write five distinct, concise descriptions of the same pose transition. Structure your descriptions as follows: "Description 1: [description]." "Description 2: [description]." \\
Ensure your descriptions accurately capture the exact pose transition without altering it. To create variety, diversify your phrasing and avoid repeating expressions. You may incorporate analogies, such as "lower your right arm to your side as if you're holding a cane," as long as the described pose transition remains unchanged.
\end{tcolorbox}
\caption{The prompt for integrating and diversifying body part-based pose transition descriptions.}
\label{fig:suppl-prompts-intdiv}
\end{figure*}

\begin{figure*}
\centering
\begin{tcolorbox}[title=Modified Prompt-2]
You are a skilled fitness\textbackslash dance instructor specializing in guiding individuals through pose transitions. \\
You will receive an image showing two poses side by side: the starting pose on the left and the target pose on the right. \\ 
Your task is to describe the transition from the starting pose to the target pose. \\
\textcolor{red}{Provide a clear and concise description of the transition.} Structure your description as follows: "Description: [description]." \\
Below are some important guidelines that must be followed:
\begin{itemize}[label=-]
\item Omit subtle or nearly imperceptible movements.
\item If the person is facing the camera, the left hand of the person is defined as the hand on your (the viewer's) right side.
\item Provide only the requested descriptions of transitions, with no additional information before or after.
\end{itemize}
\end{tcolorbox}
\caption{The modified Prompt-2 used in the ablation study for Stage \uppercase\expandafter{\romannumeral 1}. The main modification, which removes body part-specific queries, is highlighted in \textcolor{red}{red}.}
\label{fig:suppl-prompts-no-parts}
\end{figure*}

\begin{figure*}
\centering
\begin{tcolorbox}[title=Modified Prompt-3]
\textcolor{red}{You will be given a description of a specific human pose transition.} \\
Your task is to write five distinct, concise descriptions of the same pose transition. Structure your descriptions as follows: "Description 1: [description]." "Description 2: [description]." \\
Ensure your descriptions accurately capture the exact pose transition without altering it. To create variety, diversify your phrasing and avoid repeating expressions. You may incorporate analogies, such as "lower your right arm to your side as if you're holding a cane," as long as the described pose transition remains unchanged.
\end{tcolorbox}
\caption{The modified Prompt-3 used in the ablation study for Stage \uppercase\expandafter{\romannumeral 1}. The main modification, which removes body part-specific queries, is highlighted in \textcolor{red}{red}.}
\label{fig:suppl-prompts-no-parts-intdiv}
\end{figure*}

{
    \small
    \bibliographystyle{ieeenat_fullname}
    \bibliography{main}

\begin{thebibliography}{53}
\providecommand{\natexlab}[1]{#1}
\providecommand{\url}[1]{\texttt{#1}}
\expandafter\ifx\csname urlstyle\endcsname\relax
  \providecommand{\doi}[1]{doi: #1}\else
  \providecommand{\doi}{doi: \begingroup \urlstyle{rm}\Url}\fi

\bibitem[5(2025)]{unreal}
Unreal~Engine 5.
\newblock \url{https://www.unrealengine.com/en-US}, 2025.

\bibitem[Bai et~al.(2024)Bai, Wang, Xiao, He, Han, Zhang, and Shou]{bai2024hallucination}
Zechen Bai, Pichao Wang, Tianjun Xiao, Tong He, Zongbo Han, Zheng Zhang, and Mike~Zheng Shou.
\newblock Hallucination of multimodal large language models: A survey.
\newblock \emph{arXiv preprint arXiv:2404.18930}, 2024.

\bibitem[Baldrati et~al.(2023{\natexlab{a}})Baldrati, Bertini, Uricchio, and Del~Bimbo]{baldrati2023composed}
Alberto Baldrati, Marco Bertini, Tiberio Uricchio, and Alberto Del~Bimbo.
\newblock Composed image retrieval using contrastive learning and task-oriented clip-based features.
\newblock \emph{ACM Transactions on Multimedia Computing, Communications and Applications}, 20\penalty0 (3):\penalty0 1--24, 2023{\natexlab{a}}.

\bibitem[Baldrati et~al.(2023{\natexlab{b}})Baldrati, Cornia, Baraldi, and Cucchiara]{baldrati2023zero}
Alberto Baldrati, Marcella Cornia, Lorenzo Baraldi, and Rita Cucchiara.
\newblock Zero-shot composed image retrieval with textual inversion.
\newblock In \emph{CVPR}, 2023{\natexlab{b}}.

\bibitem[Black et~al.(2023)Black, Patel, Tesch, and Yang]{Black_CVPR_2023}
Michael~J. Black, Priyanka Patel, Joachim Tesch, and Jinlong Yang.
\newblock {BEDLAM}: A synthetic dataset of bodies exhibiting detailed lifelike animated motion.
\newblock In \emph{Proceedings IEEE/CVF Conf.~on Computer Vision and Pattern Recognition (CVPR)}, pages 8726--8737, 2023.

\bibitem[Cao et~al.(2017)Cao, Hidalgo, Simon, Wei, and Sheikh]{cao2017realtime}
Zhe Cao, Gines Hidalgo, Tomas Simon, Shih-En Wei, and Yaser Sheikh.
\newblock Realtime multi-person 2d pose estimation using part affinity fields.
\newblock In \emph{CVPR}, 2017.

\bibitem[Chen et~al.(2024{\natexlab{a}})Chen, Ye, He, Wang, Khashabi, and Yuille]{chen2024efficient}
Jieneng Chen, Luoxin Ye, Ju He, Zhao-Yang Wang, Daniel Khashabi, and Alan Yuille.
\newblock Efficient large multi-modal models via visual context compression.
\newblock In \emph{NEURIPS}, 2024{\natexlab{a}}.

\bibitem[Chen et~al.(2024{\natexlab{b}})Chen, Qin, Jiang, and Choi]{chen2024large}
Ruirui Chen, Chengwei Qin, Weifeng Jiang, and Dongkyu Choi.
\newblock Is a large language model a good annotator for event extraction?
\newblock In \emph{AAAI}, 2024{\natexlab{b}}.

\bibitem[Delmas et~al.(2022)Delmas, Weinzaepfel, Lucas, Moreno-Noguer, and Rogez]{delmas2022posescript}
Ginger Delmas, Philippe Weinzaepfel, Thomas Lucas, Francesc Moreno-Noguer, and Grégory Rogez.
\newblock Posescript: 3d human poses from natural language.
\newblock In \emph{ECCV}, 2022.

\bibitem[Delmas et~al.(2023)Delmas, Weinzaepfel, Moreno-Noguer, and Rogez]{delmas2023posefix}
Ginger Delmas, Philippe Weinzaepfel, Francesc Moreno-Noguer, and Gr{\'e}gory Rogez.
\newblock Posefix: correcting 3d human poses with natural language.
\newblock In \emph{Proceedings of the IEEE/CVF International Conference on Computer Vision}, pages 15018--15028, 2023.

\bibitem[Feng et~al.(2024)Feng, Lin, Dwivedi, Sun, Patel, and Black]{feng2024chatpose}
Yao Feng, Jing Lin, Sai~Kumar Dwivedi, Yu Sun, Priyanka Patel, and Michael~J. Black.
\newblock Chatpose: Chatting about 3d human pose.
\newblock In \emph{Proceedings of the IEEE/CVF Conference on Computer Vision and Pattern Recognition (CVPR)}, 2024.

\bibitem[Ferrari et~al.(2009)Ferrari, Marin-Jimenez, and Zisserman]{ferrari2009pose}
Vittorio Ferrari, Manuel Marin-Jimenez, and Andrew Zisserman.
\newblock Pose search: Retrieving people using their pose.
\newblock In \emph{CVPR}, 2009.

\bibitem[G{\"u}ler et~al.(2018)G{\"u}ler, Neverova, and Kokkinos]{guler2018densepose}
Riza~Alp G{\"u}ler, Natalia Neverova, and Iasonas Kokkinos.
\newblock Densepose: Dense human pose estimation in the wild.
\newblock In \emph{CVPR}, 2018.

\bibitem[Haven(2025)]{hdri}
Poly Haven.
\newblock \url{https://polyhaven.com/hdris}, 2025.

\bibitem[Hoffman et~al.(2018)Hoffman, Tzeng, Park, Zhu, Isola, Saenko, Efros, and Darrell]{hoffman2018cycada}
Judy Hoffman, Eric Tzeng, Taesung Park, Jun-Yan Zhu, Phillip Isola, Kate Saenko, Alexei Efros, and Trevor Darrell.
\newblock Cycada: Cycle-consistent adversarial domain adaptation.
\newblock In \emph{International conference on machine learning}, pages 1989--1998. Pmlr, 2018.

\bibitem[Hu et~al.(2022)Hu, Shen, Wallis, Allen-Zhu, Li, Wang, Wang, Chen, et~al.]{hu2022lora}
Edward~J Hu, Yelong Shen, Phillip Wallis, Zeyuan Allen-Zhu, Yuanzhi Li, Shean Wang, Lu Wang, Weizhu Chen, et~al.
\newblock Lora: Low-rank adaptation of large language models.
\newblock \emph{ICLR}, 1\penalty0 (2):\penalty0 3, 2022.

\bibitem[Hurst et~al.(2024)Hurst, Lerer, Goucher, Perelman, Ramesh, Clark, Ostrow, Welihinda, Hayes, Radford, et~al.]{hurst2024gpt}
Aaron Hurst, Adam Lerer, Adam~P Goucher, Adam Perelman, Aditya Ramesh, Aidan Clark, AJ Ostrow, Akila Welihinda, Alan Hayes, Alec Radford, et~al.
\newblock Gpt-4o system card.
\newblock \emph{arXiv preprint arXiv:2410.21276}, 2024.

\bibitem[Jang et~al.(2024)Jang, Kim, Meng, Huynh, and Lim]{jang2024visual}
Young~Kyun Jang, Donghyun Kim, Zihang Meng, Dat Huynh, and Ser-Nam Lim.
\newblock Visual delta generator with large multi-modal models for semi-supervised composed image retrieval.
\newblock In \emph{Proceedings of the IEEE/CVF Conference on Computer Vision and Pattern Recognition}, pages 16805--16814, 2024.

\bibitem[Jenicek and Chum(2019)]{jenicek2019linking}
Tomas Jenicek and Ond{\v{r}}ej Chum.
\newblock Linking art through human poses.
\newblock In \emph{2019 International Conference on Document Analysis and Recognition (ICDAR)}, pages 1338--1345. IEEE, 2019.

\bibitem[Kim et~al.(2021)Kim, Yu, Kim, and Kim]{kim2021fixmypose}
Jongseok Kim, Youngjae Yu, Hoeseong Kim, and Gunhee Kim.
\newblock Dual compositional learning in interactive image retrieval.
\newblock In \emph{AAAI}, pages 1771--1779, 2021.

\bibitem[Lee et~al.(2021)Lee, Kim, and Han]{lee2021cosmo}
Seungmin Lee, Dongwan Kim, and Bohyung Han.
\newblock Cosmo: Content-style modulation for image retrieval with text feedback.
\newblock In \emph{Proceedings of the IEEE/CVF Conference on Computer Vision and Pattern Recognition}, pages 802--812, 2021.

\bibitem[Li et~al.(2022)Li, Li, Xiong, and Hoi]{li2022blip}
Junnan Li, Dongxu Li, Caiming Xiong, and Steven Hoi.
\newblock Blip: Bootstrapping language-image pre-training for unified vision-language understanding and generation.
\newblock In \emph{International conference on machine learning}, pages 12888--12900. PMLR, 2022.

\bibitem[Li et~al.(2021)Li, Yang, Ross, and Kanazawa]{li2021ai}
Ruilong Li, Shan Yang, David~A Ross, and Angjoo Kanazawa.
\newblock Ai choreographer: Music conditioned 3d dance generation with aist++.
\newblock In \emph{Proceedings of the IEEE/CVF international conference on computer vision}, pages 13401--13412, 2021.

\bibitem[Liu et~al.(2024)Liu, Li, Li, and Lee]{liu2024improved}
Haotian Liu, Chunyuan Li, Yuheng Li, and Yong~Jae Lee.
\newblock Improved baselines with visual instruction tuning.
\newblock In \emph{Proceedings of the IEEE/CVF Conference on Computer Vision and Pattern Recognition}, pages 26296--26306, 2024.

\bibitem[Liu et~al.(2021{\natexlab{a}})Liu, Rodriguez-Opazo, Teney, and Gould]{liu2021cirr}
Zheyuan Liu, Cristian Rodriguez-Opazo, Damien Teney, and Stephen Gould.
\newblock Image retrieval on real-life images with pre-trained vision-and-language models.
\newblock In \emph{ICCV}, pages 2125--2134, 2021{\natexlab{a}}.

\bibitem[Liu et~al.(2021{\natexlab{b}})Liu, Rodriguez-Opazo, Teney, and Gould]{liu2021image}
Zheyuan Liu, Cristian Rodriguez-Opazo, Damien Teney, and Stephen Gould.
\newblock Image retrieval on real-life images with pre-trained vision-and-language models.
\newblock In \emph{ICCV}, pages 2125--2134, 2021{\natexlab{b}}.

\bibitem[Loper et~al.(2015)Loper, Mahmood, Romero, Pons-Moll, and Black]{SMPL:2015}
Matthew Loper, Naureen Mahmood, Javier Romero, Gerard Pons-Moll, and Michael~J. Black.
\newblock {SMPL}: A skinned multi-person linear model.
\newblock \emph{ACM Trans. Graphics (Proc. SIGGRAPH Asia)}, 34\penalty0 (6):\penalty0 248:1--248:16, 2015.

\bibitem[Mahmood et~al.(2019)Mahmood, Ghorbani, Troje, Pons-Moll, and Black]{mahmood2019amass}
Naureen Mahmood, Nima Ghorbani, Nikolaus~F Troje, Gerard Pons-Moll, and Michael~J Black.
\newblock Amass: Archive of motion capture as surface shapes.
\newblock In \emph{Proceedings of the IEEE/CVF international conference on computer vision}, pages 5442--5451, 2019.

\bibitem[Meshcapade~GmbH(2025)]{meshcapade}
Germany Meshcapade~GmbH, T¨~ubingen.
\newblock \url{https://meshcapade.com/}, 2025.

\bibitem[Mori et~al.(2015)Mori, Pantofaru, Kothari, Leung, Toderici, Toshev, and Yang]{mori2015pose}
Greg Mori, Caroline Pantofaru, Nisarg Kothari, Thomas Leung, George Toderici, Alexander Toshev, and Weilong Yang.
\newblock Pose embeddings: A deep architecture for learning to match human poses, 2015.

\bibitem[Patel et~al.(2021)Patel, Huang, Tesch, Hoffmann, Tripathi, and Black]{Patel:CVPR:2021}
Priyanka Patel, Chun-Hao~P. Huang, Joachim Tesch, David~T. Hoffmann, Shashank Tripathi, and Michael~J. Black.
\newblock {AGORA}: Avatars in geography optimized for regression analysis.
\newblock In \emph{Proceedings IEEE/CVF Conf.~on Computer Vision and Pattern Recognition ({CVPR})}, 2021.

\bibitem[Pavlakos et~al.(2018)Pavlakos, Zhou, and Daniilidis]{pavlakos2018ordinal}
Georgios Pavlakos, Xiaowei Zhou, and Kostas Daniilidis.
\newblock Ordinal depth supervision for 3d human pose estimation.
\newblock In \emph{CVPR}, 2018.

\bibitem[Pavlakos et~al.(2019)Pavlakos, Choutas, Ghorbani, Bolkart, Osman, Tzionas, and Black]{SMPL-X:2019}
Georgios Pavlakos, Vasileios Choutas, Nima Ghorbani, Timo Bolkart, Ahmed A.~A. Osman, Dimitrios Tzionas, and Michael~J. Black.
\newblock Expressive body capture: {3D} hands, face, and body from a single image.
\newblock In \emph{Proceedings IEEE Conf. on Computer Vision and Pattern Recognition (CVPR)}, pages 10975--10985, 2019.

\bibitem[Petrovich et~al.(2021)Petrovich, Black, and Varol]{petrovich2021action}
Mathis Petrovich, Michael Black, and G{\"u}l Varol.
\newblock Action-conditioned 3d human motion synthesis with transformer vae.
\newblock In \emph{ICCV}, 2021.

\bibitem[Radford et~al.(2021)Radford, Kim, Hallacy, Ramesh, Goh, Agarwal, Sastry, Askell, Mishkin, Clark, Krueger, and Sutskever]{radford2021learning}
Alec Radford, Jong~Wook Kim, Chris Hallacy, Aditya Ramesh, Gabriel Goh, Sandhini Agarwal, Girish Sastry, Amanda Askell, Pamela Mishkin, Jack Clark, Gretchen Krueger, and Ilya Sutskever.
\newblock Learning transferable visual models from natural language supervision.
\newblock In \emph{Proceedings of the International Conference on Machine Learning (ICML)}, 2021.

\bibitem[Saito et~al.(2023)Saito, Sohn, Zhang, Li, Lee, Saenko, and Pfister]{saito2023pic2word}
Kuniaki Saito, Kihyuk Sohn, Xiang Zhang, Chun-Liang Li, Chen-Yu Lee, Kate Saenko, and Tomas Pfister.
\newblock Pic2word: Mapping pictures to words for zero-shot composed image retrieval.
\newblock In \emph{Proceedings of the IEEE/CVF Conference on Computer Vision and Pattern Recognition}, pages 19305--19314, 2023.

\bibitem[Shen et~al.(2024)Shen, Lee, Kwon, and Bhattacharyya]{shen2024diversifying}
Yi-Ting Shen, Hyungtae Lee, Heesung Kwon, and Shuvra~S Bhattacharyya.
\newblock Diversifying human pose in synthetic data for aerial-view human detection.
\newblock \emph{arXiv preprint arXiv:2405.15939}, 2024.

\bibitem[Shorten and Khoshgoftaar(2019)]{shorten2019survey}
Connor Shorten and Taghi~M. Khoshgoftaar.
\newblock A survey on image data augmentation for deep learning.
\newblock \emph{Journal of Big Data}, 6\penalty0 (1):\penalty0 1--48, 2019.

\bibitem[Song et~al.(2024)Song, Lin, Wen, Hou, Xu, and Nie]{song2024survey}
Xuemeng Song, Haoqiang Lin, Haokun Wen, Bohan Hou, Mingzhu Xu, and Liqiang Nie.
\newblock A comprehensive survey on composed image retrieval.
\newblock \emph{ACM Transactions on Information Systems}, 1\penalty0 (1):\penalty0 1--45, 2024.

\bibitem[Su et~al.(2023)Su, Kasai, Wu, Shi, Wang, Xin, Zhang, Ostendorf, Zettlemoyer, Smith, and Yu]{su2023selective}
Hongjin Su, Jungo Kasai, Chen~Henry Wu, Weijia Shi, Tianlu Wang, Jiayi Xin, Rui Zhang, Mari Ostendorf, Luke Zettlemoyer, Noah~A. Smith, and Tao Yu.
\newblock Selective annotation makes language models better few-shot learners.
\newblock In \emph{ICLR}, 2023.

\bibitem[Tan et~al.(2024)Tan, Beigi, Wang, Guo, Bhattacharjee, Jiang, Karami, Li, Cheng, and Liu]{tan2024large}
Zhen Tan, Alimohammad Beigi, Song Wang, Ruocheng Guo, Amrita Bhattacharjee, Bohan Jiang, Mansooreh Karami, Jundong Li, Lu Cheng, and Huan Liu.
\newblock Large language models for data annotation and synthesis: A survey.
\newblock \emph{EMNLP}, 2024.

\bibitem[Vo et~al.(2019)Vo, Jiang, Sun, Murphy, Li, Fei-Fei, and Hays]{vo2019composing}
Nam Vo, Lu Jiang, Chen Sun, Kevin Murphy, Li-Jia Li, Li Fei-Fei, and James Hays.
\newblock Composing text and image for image retrieval-an empirical odyssey.
\newblock In \emph{Proceedings of the IEEE/CVF conference on computer vision and pattern recognition}, pages 6439--6448, 2019.

\bibitem[Wang et~al.(2024)Wang, Yu, Li, Jia, and Ding]{wang2024adaptive}
Ge Wang, Fei Yu, Jian Li, Qi Jia, and Shouhong Ding.
\newblock Adaptive uncertainty-based learning for text-based person retrieval.
\newblock In \emph{AAAI}, 2024.

\bibitem[Wang et~al.(2018)Wang, Guo, Yan, Quan, and Zhang]{wang2018learning}
Hanyu Wang, Junhui Guo, Deyou~Ma Yan, Weimin Quan, and Xinge Zhang.
\newblock Learning 3d keypoint descriptors for non-rigid shape matching.
\newblock In \emph{ECCV}, 2018.

\bibitem[Wen et~al.(2024)Wen, Song, Chen, Wei, Nie, and Chua]{wen2024fusion}
Haokun Wen, Xuemeng Song, Xiaolin Chen, Yinwei Wei, Liqiang Nie, and Tat-Seng Chua.
\newblock Simple but effective raw-data level multimodal fusion for composed image retrieval.
\newblock In \emph{Proceedings of the International ACM SIGIR Conference on Research and Development in Information Retrieval}, pages 229--239, 2024.

\bibitem[Wu et~al.(2021)Wu, Gao, Guo, Al-Halah, Rennie, Grauman, and Feris]{wu2021fashion}
Hui Wu, Yupeng Gao, Xiaoxiao Guo, Ziad Al-Halah, Steven Rennie, Kristen Grauman, and Rogerio Feris.
\newblock Fashion iq: A new dataset towards retrieving images by natural language feedback.
\newblock In \emph{CVPR}, pages 11307--11317, 2021.

\bibitem[Wu et~al.(2023)Wu, Gan, Chen, Wan, and Philip]{wu2023multimodal}
Jiayang Wu, Wensheng Gan, Zefeng Chen, Shicheng Wan, and S~Yu Philip.
\newblock Multimodal large language models: A survey.
\newblock In \emph{2023 IEEE International Conference on Big Data (BigData)}, pages 2247--2256. IEEE, 2023.

\bibitem[Wu et~al.(2024{\natexlab{a}})Wu, Zhong, Xing, Lai, Liu, Wang, Chen, Zhu, Lu, Luo, et~al.]{wu2024visionllm}
Jiannan Wu, Muyan Zhong, Sen Xing, Zeqiang Lai, Zhaoyang Liu, Wenhai Wang, Zhe Chen, Xizhou Zhu, Lewei Lu, Ping Luo, et~al.
\newblock Visionllm v2: An end-to-end generalist multimodal large language model for hundreds of vision-language tasks.
\newblock \emph{NEURIPS}, 2024{\natexlab{a}}.

\bibitem[Wu et~al.(2024{\natexlab{b}})Wu, Fei, Qu, Ji, and Chua]{wu2024next}
Shengqiong Wu, Hao Fei, Leigang Qu, Wei Ji, and Tat-Seng Chua.
\newblock Next-gpt: Any-to-any multimodal llm.
\newblock In \emph{Forty-first International Conference on Machine Learning}, 2024{\natexlab{b}}.

\bibitem[Xu et~al.(2022)Xu, Zhang, Zhang, and Tao]{xu2022vitpose}
Yufei Xu, Jing Zhang, Qiming Zhang, and Dacheng Tao.
\newblock Vitpose: Simple vision transformer baselines for human pose estimation.
\newblock \emph{Advances in neural information processing systems}, 35:\penalty0 38571--38584, 2022.

\bibitem[Zhang et~al.(2023)Zhang, Huang, and Wang]{zhang2023motiongpt}
Yifan Zhang, Cheng Huang, and Limin Wang.
\newblock Motiongpt: Large language model for human motion generation.
\newblock In \emph{Advances in Neural Information Processing Systems (NeurIPS)}, 2023.

\bibitem[Zhao and Xu(2024)]{zhao2024neucore}
Shu Zhao and Huijuan Xu.
\newblock Neucore: Neural concept reasoning for composed image retrieval.
\newblock In \emph{Proceedings of UniReps: the First Workshop on Unifying Representations in Neural Models}, pages 47--59. PMLR, 2024.

\bibitem[Zhu et~al.(2017)Zhu, Park, Isola, and Efros]{zhu2017unpaired}
Jun-Yan Zhu, Taesung Park, Phillip Isola, and Alexei~A. Efros.
\newblock Unpaired image-to-image translation using cycle-consistent adversarial networks.
\newblock In \emph{Proceedings of the IEEE International Conference on Computer Vision (ICCV)}, 2017.

\end{thebibliography}
}

\end{document}